\documentclass[10pt,journal,final,twocolumn]{IEEEtran}

\usepackage{graphicx,times,amsmath,amssymb,cite,cases,mathrsfs,booktabs,multirow,makecell,psfrag,subfigure,url,stfloats}
\usepackage{algorithm,algorithmic,amsthm,ulem,latexsym,bm}
\usepackage{dsfont}
\usepackage[colorlinks,linkcolor=red,anchorcolor=blue,citecolor=red]{hyperref}

\newtheorem*{definition}{Definition}

\begin{document}
\title{Discriminative Residual Analysis for Image Set Classification with Posture and Age Variations}

\author{Chuan-Xian~Ren, You-Wei Luo, Xiao-Lin~Xu, Dao-Qing~Dai, Hong~Yan~\IEEEmembership{Fellow,~IEEE}
\thanks{C.X. Ren, Y.W. Luo, and D.Q. Dai are with the Intelligent Data Center, School of Mathematics, Sun Yat-Sen University, Guangzhou, China.}
\thanks{X.L. Xu is with the School of Statistics and Mathematics, Guangdong University of Finance and Economics, Guangzhou, China.}
\thanks{H. Yan is with the Department of Electronic Engineering, City University of Hong Kong, 83 Tat Chee Avenue, Kowloon, Hong Kong.}
\thanks{This work is supported in part by the National Natural Science Foundation of China under Grants 61572536, 11631015, and U1611265, in part by the Science and Technology Program of Guangzhou under Grant 201804010248, and in part by the Hong Kong Research Grants Council (Project C1007-15G).}}


\graphicspath{{images/}}
\maketitle

\begin{abstract}
Image set recognition has been widely applied in many practical problems like real-time video retrieval and image caption tasks. Due to its superior performance, it has grown into a significant topic in recent years. However, images with complicated variations, e.g., postures and human ages, are difficult to address, as these variations are continuous and gradual with respect to image appearance. Consequently, the crucial point of image set recognition is to mine the intrinsic connection or structural information from the image batches with variations. In this work, a Discriminant Residual Analysis (DRA) method is proposed to improve the classification performance by discovering discriminant features in related and unrelated groups. Specifically, DRA attempts to obtain a powerful projection which casts the \textit{residual} representations into a discriminant subspace. Such a projection subspace is expected to magnify the useful information of the input space as much as possible, then the relation between the training set and the test set described by the given metric or distance will be more precise in the discriminant subspace. We also propose a nonfeasance strategy by defining another approach to construct the unrelated groups, which help to reduce furthermore the cost of sampling errors. Two regularization approaches are used to deal with the probable small sample size problem. Extensive experiments are conducted on benchmark databases, and the results show superiority and efficiency of the new methods.
\end{abstract}


\IEEEpeerreviewmaketitle

\section{Introduction}\label{sec1}

\begin{figure}[t]
\begin{center}
\includegraphics[width=1.0\linewidth]{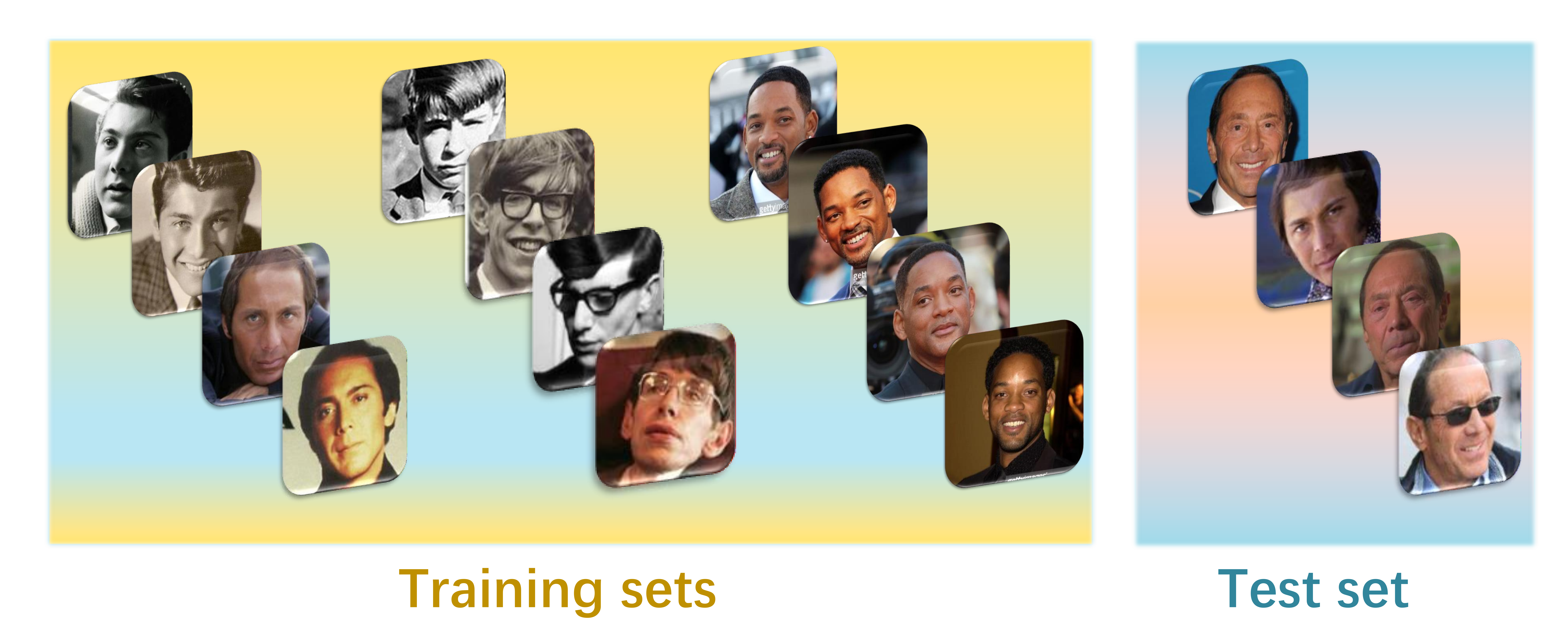}
\end{center}
\vspace{-1pc}
\caption{Demonstration of the image set classification problem. The test set to be labeled is a set, rather than just one sample. Meanwhile, there is a large variation within each image set.}
\label{fig:problem}
\end{figure}

\IEEEPARstart{I}{mage} set recognition is an important issue in computer vision and pattern recognition, as it has wide applications such as video retrieval and image caption \cite{MMDML,wang2017discriminative,kim2007ImageSet,huang2015log,wang2018discriminant,zhao19review}. Unlike classical methods, such as Support Vector Machine (SVM) \cite{SVM} and Collaborative Representation based Classifier (CRC) \cite{CRC}, image set recognition performs batch/set verification or identification on the training set and test set. As shown in Fig.~\ref{fig:problem}, the training set consists of several classes including \textit{Anka}, \textit{Hawking} and \textit{Smith}, while the test set is a group of \textit{Anka} samples, rather than individual samples having different labels. The target is to predict class label of the coming test set. In this case, many traditional methods designed just for the single image classification tasks, such as SVM and CRC will be unsuitable any more.

Another challenge of image set recognition is that some complex imaging variations (e.g., posture, age, and light) are difficult to characterize, refine and deal with. In many cases, complex environmental changes cause the intra-class variance to be much larger than the inter-class variance~\cite{kim2007ImageSet,wang2018discriminant,ren2016enhanced}. It means that there possibly be a large overlap between different classes, which will easily lead to misjudgment. In particular, these image variations are continuous, gradual and subtle with respect to image appearances, so it is difficult to extract discriminant information for classification. The idea of image set recognition can be used to extract discriminant features across variation species.

An important goal of image set recognition is to learn the continuity information, e.g., facial manifold in different illuminations, postures and expressions, as shown in Fig.~\ref{fig:problem}. The crucial point of image recognition is then cast to mining the intrinsic connection or structural information from the image batches with different categories. Recently, several approaches based on representation learning or dictionary learning are proposed for efficient image set recognition \cite{wang2018discriminant,wang2017discriminative,SFDL,DMK,ren2016enhanced}. These methods can capture the structured information of images and preserve it in features or a dictionary. Especially, Discriminant Analysis on Riemannian Manifold of Gaussian Distributions (DARG) \cite{wang2018discriminant} models each image set with a Gaussian mixture model and learns the discriminative information from high-dimensional Hilbert space. Other strategies try to extract discriminant information from the raw sample space, which helps to enhance the prediction accuracy \cite{Yao1,DGK}. To exploit intrinsic connection and joint features, Zheng et al. \cite{zheng2017Sparse} propose to learn the extended cooperative sparse representation for both training set and test set. In \cite{MMDML} and \cite{DML}, the authors attempt to explore a significant metric for image set, which focuses on simultaneously optimizing the within-class similarity and between-class diversity. In \cite{wang2017prototype}, Prototype Discriminative Learning (PDL) is proposed to search the virtual prototypes of raw images and learn the linear discriminative projection of prototypes simultaneously.

With the rapid advances in deep learning literature \cite{ren2018generalized,DNN,yang2017neural,sohn2017adv,shah2016iterative}, convolutional neural networks (CNN) offer another learning framework for image set recognition. Yang et al. \cite{yang2017neural} propose the Neural Aggregation Network (NAN) method, which employs deep CNN as a frame-level feature extractor and provides an unsupervised technique to learn a weighted combination of all frame-level features. Following NAN, Sohn et al. \cite{sohn2017adv} deal with the unlabeled videos face recognition by transferring the discriminant features from the labeled images (source domain) to the unlabeled videos (target domain), where NAN facilitates the discriminant features extraction. Based on the deep learning mechanism, Shah et al. \cite{shah2016iterative} present an Iterative Deep Learning Model (IDLM) to hierarchically learn class-specific image set representations. It preserves invariant information at lower levels and learns discriminant features at higher levels. In practice, most of these methods require a large amount of data by default, which limits their further applications.

Linear regression models have been extended to address the image set classification problem \cite{wang2008ManifoldMD}. Dual Linear Regression Classification (DLRC) \cite{DLRC} defines the \textit{virtual} appearance space to exploit the relationship between the training set and the test set explicitly. Recently, Pairwise Linear Regression Classification (PLRC) \cite{PLRC} extends DLRC by constructing both related groups and unrelated groups, which distinguish the neighborhoods of given instances by specified distance and their category information. By using the defined groups, both DLRC and PLRC establish new procedures to predict labels of the test sets. However, these reconstruction-based methods cannot exploit intrinsic connection between different image sets and learn discriminant features for classification.

The motivation of our method can be described as follows. On one hand, though DLRC and PLRC construct the virtual appearance space, there is still much redundant and even noisy information exists in the raw data space, which will mislead the classifier and produce unfavorable prediction results. To effectively extract more discriminative features from such a messy space, the influence of noisy variations must be minimized. As the residual space lightens the negative effects from the variations and aims to preserve the task-specific features, it is more suitable for the discriminant learning. On the other hand, several classifiers have inadequate generalization performance when the image sets are of small sizes, which restricts the application range of classification algorithms in practical scenarios. So it is meaningful to address this \textit{small sample size} problem, which is also discussed in \cite{DLRC,ren2015sample}. Excellent classification methods should be able to deal with such recognition tasks efficiently and accurately.

In this paper, we propose a novel Discriminant Residual Analysis (DRA) method to deal with these problems. We firstly define the \textit{distance of interest}, which is an important criterion of subspace information screening in DRA. To capture the interested information in \textit{residual} representation space, DRA tries to find a powerful map which projects the \textit{residual} representations into a discriminant subspace. In such a subspace, the useful information is magnified; then the positive set pairs (i.e., a pair of sets from the same class) are closer to each other and the negative set pairs (i.e., a pair of sets from different classes) are more dispersed. Even with a few number of training and validation samples, DRA is still effective in maintaining a high level of recognition ability. Besides, a nonfeasance strategy (NFS) is proposed to redefine the unrelated groups by removing the distance metric and skipping the samples selection model for unrelated groups in PLRC. Alternatively, NFS exploits another way to form the unrelated samples. In this case, the incorrectly sample selection is avoided when constructing the unrelated groups and the classification performance is improved.

Our contributions are summarized as follows:
\begin{itemize}
	\item We propose a novel discriminant residual learning algorithm for image set recognition. In contrary to conventional discriminant analysis, the new method is built on a residual space, rather than the appearance space. The interested residuals will be mined and then used to extract discriminant features.
	\item We propose NFS to redefine the unrelated groups in our framework. NFS avoids the occurrence of inappropriate samples selection when forming unrelated groups and exhibits better classification performance.
    \item The DRA method automatically learns discriminant information from the residual representation space, so it does not rely on any geometric assumptions. Experiment results show the superiority of the DRA method.
\end{itemize}

The rest of this paper is organized as follows. In Section \ref{sec2}, we briefly review the recent development of image set based recognition methods. In Section \ref{sec3}, we define the problem and settings of image set recognition, and then present the DRA and NFS methods. The experiment results of our methods are shown and then compared with other state-of-the-art methods in Section \ref{sec4}. Section \ref{sec5} concludes the paper.

\section{Related Work}\label{sec2}

In this section, we briefly review the image set based recognition literature and summarize some advanced approaches.

Linear subspace learning is a classic and simple way for efficient image set recognition \cite{kim2007ImageSet}. It assumes that each image set represents a linear space. Then the similarity measurement will be used to reflect the correlations between different sets. In order to extend it to a nonlinear or manifold subspace, Cevikalp et al. \cite{HISD} develop the Affine Hull based Image Set Distance (AHISD) and Convex Hull based Image Set Distance (CHISD), which employs affine/convex hull transformation to represent image sets are as points in affine/convex subspace. Motivated by the hull based methods, Zhu et al. \cite{RISCRC} proposes the image set based collaborative representation and classification (ISCRC) approach to represent the hull collaboratively over all the gallery sets. Wang et al. \cite{MMDML,wang2008ManifoldMD} propose the Manifold-Manifold Distance which describes each image set as a manifold subspace.

Metric learning methods have been developed to learn a suitable distance measurement for image set \cite{DML,sohn2017adv} in recent years. Under such metrics, distances between image sets should be as small as possible for the homogeneous sets and as large as possible for the heterogeneous sets. Compared with the pairwise constraints, triplet constraints serve the purpose of reducing the computational complexity \cite{huang2015log,qian2015fine}.

Representation learning or dictionary learning methods attempt to find the structures or key features of image sets \cite{wang2018discriminant,wang2017discriminative,SFDL,DMK,zheng2017Sparse,shah2016iterative}. To some extent, the feature representation should maximize discriminant ability or minimize the reconstruction error. In general, the main idea of most image set feature learning approaches are to make the inter-classes more separable and the intra-classes more compact \cite{ren2016enhanced,Yao1,DGK,yu2018kernel}. Moreover, feature extractors based on CNNs show higher accuracies than traditional hand-crafted methods \cite{ren2018generalized,yang2017neural,sohn2017adv,ren2019deep}. However, neural networks are black boxes and they are lack of adequate interpretations.

We also notice that the unconstrained images recognition has attracted widespread attention. A wide variety of databases have also been collected and made public in the past decade \cite{agedb}. Labeled Faces ``in-the-wild'' (LFW) \cite{LFW} is one of the earliest \textit{in-the-wild} databases. Compared with the controlled conditions (e.g., illuminations, poses, expressions, etc.) in conventional data sets, \textit{in-the-wild} databases contain images with large variations in backgrounds, age, appearance, pixel, style, occlusion and so on. Generally, the uncontrolled images can be obtained in two main ways. The first one consists of web page images of celebrities, such as CelebA~\cite{CelebA}, MegaFace~\cite{megaface} and VggFace2~\cite{VGGface2}. Most of them contain massive images which are obtained by combining automatic search and manual screening. Another way is to collect video frame images like YTF~\cite{YTF} and IJB-A~\cite{IJB-A}, since the position and shape of objects or humans in videos are uncontrolled. Video frames often contain a lot of repetitive but continuous information, which is the key issue to learn the structures or features of video image sets.

Our work is to learn a discriminant subspace of \textit{residual} representations from image sets, and further improve the performance of image set based method. Moreover, we hope that the proposed method can adapt well to the complicated real scenarios, e.g., the \textit{in-the-wild} conditions.

\section{Discriminant Residual Analysis}\label{sec3}

In this section, we present our motivation and define the learning problem in Section \ref{sec3.1}. Then the DRA method is introduced in Section \ref{sec3.2}. Section \ref{sec3.3} provides an alternative definition approach of unrelated group, i.e., NFS. Section \ref{sec3.4} discusses the singularity problem appeared in numerical optimization, then presents two regularization approaches to address it. Section \ref{sec3.5} presents the complexity analysis.

\subsection{Problem Definition and Settings}\label{sec3.1}

By capturing the information of related groups and unrelated groups simultaneously, DLRC and PLRC exhibit efficiency and effectiveness in the image set based classification task. Given data matrix ${\bf {\bf X}}_k\in \mathbb{R}^{d\times m_k}$ with ground-truth label $k$, and the test set ${\bf Y}\in \mathbb{R}^{d\times n}$. Here $d$ denotes the dimension of image vectors and the second dimensions, i.e., $m_k$ and $n$, represent the number of samples. Mathematically, DLRC and PLRC attempt to find the virtual representation by solving the regression problem as follows:
\begin{equation}\label{eq:1}
{\bf {\bf X}}_{k}\alpha_1 = {\bf Y}\alpha_2 \triangleq {\bf V},
\end{equation}
where $\alpha_1$ and $\alpha_2$ are regression coefficients to be determined, ${\bf V}$ is the so-called virtual face \cite{DLRC,PLRC}.

\begin{figure*}[htb]
\begin{center}
\includegraphics[width=0.95\linewidth]{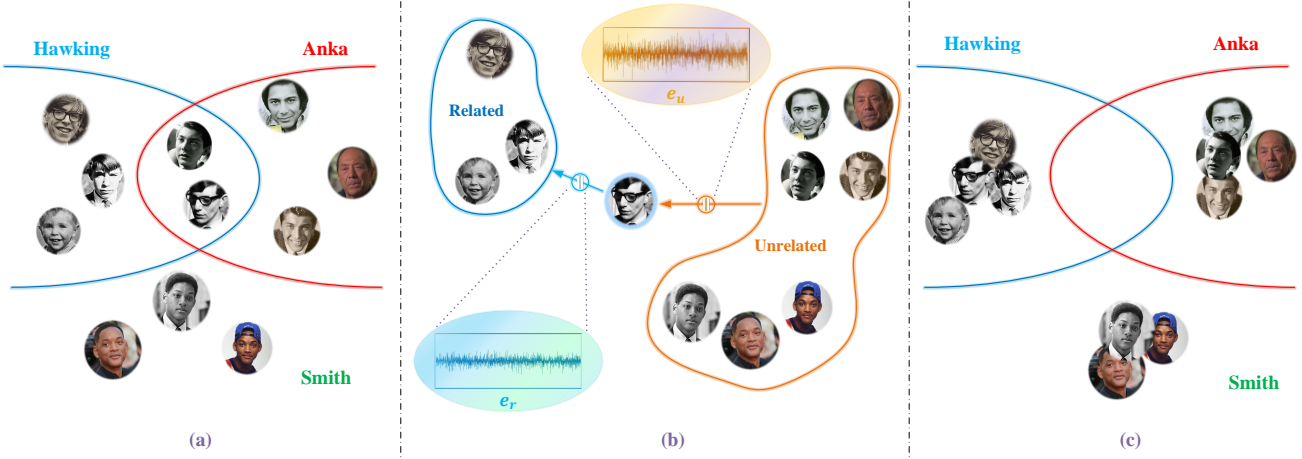}
\end{center}
\vspace{-1pc}
\caption{Flowchart of the DRA method. The subspace learning module aims to make the data in the same class more compact and data between different classes more apart. (a) Raw data space. (b) The projection operation with unrelated group construction. (c) The final embedding space. Specifically, the DOI is defined and employed to form  the \textit{residual} model, which is also taken as the regression error in raw data space.}
\label{fig:flowchart}
\end{figure*}

Given training set ${\bf {\bf X}} \triangleq [{\bf {\bf X}}_1, {\bf {\bf X}}_2, \cdots, {\bf {\bf X}}_c] \in \mathbb{R}^{d \times m}$ and test set ${\bf Y}\in \mathbb{R}^{d \times n}$ with ground-truth label $l$. The related group of the $k$-th class is ${\bf X}_k=[{\bf x}^k_1, {\bf x}^k_2, \ldots, {\bf x}^k_{m_k}]\in \mathbb{R}^{d\times m_k}$ and the unrelated group ${\bf U}_k = [{\bf u}^k_1, {\bf u}^k_2, \ldots, {\bf u}^k_{m'_k}]\in \mathbb{R}^{d\times m'_k}$ can be derived through different strategies (e.g., the distance metric in PLRC \cite{PLRC}). To measure the related distance $d_r$ and unrelated distance $d_u$, the solutions of Eq.~\eqref{eq:1} are required. In practice, we transform this equation to a more common linear regression problem by building intermediate variables as follows,
\begin{align}
\hat{{\bf X}}_k &= [{\bf x}^k_1,\ldots,{\bf x}^k_{m_k-1}] - {\bf x}^k_{m_k}{\bf 1}^T_{m_k-1},\nonumber \\
\hat{{\bf U}}_{k} &= [{\bf u}^{k}_1,\ldots,{\bf u}^{k}_{m'_k-1}] - {\bf u}^{k}_{m'_k}{\bf 1}^T_{m'_k-1} ,\nonumber \\
\hat{{\bf Y}} &= [{\bf y}_1,\ldots,{\bf y}_{n-1}] - {\bf y}_{n}{\bf 1}^T_{n-1} .\nonumber
\end{align}
Then the regression coefficients $\gamma^{k}$ and $\varepsilon^{k}$ are obtained by solving the following regression problems
\begin{equation} \label{eq:2}
  \begin{split}
  [\hat{{\bf X}}_k, -\hat{{\bf Y}}]\gamma^{k}={\bf y}_{n} - {\bf x}^k_{m_k},    \\
  [\hat{{\bf U}}_{k}, -\hat{{\bf Y}}]\varepsilon^{k}={\bf y}_{n} - {\bf u}^{k}_{m'_k}.
  \end{split}
\end{equation}
Recall that the classification metric consists of the related distance $d_r$ and unrelated distance $d_u$. Let
\begin{equation} \label{eq:3}
  \begin{split}
  {\bf e}_r^{k} &= [\hat{{\bf X}}_k, -\hat{{\bf Y}}]\gamma^{k} - ({\bf y}_{n} - {\bf x}^k_{m_k}),    \\
  {\bf e}_u^{k} &= [\hat{{\bf U}}_k, -\hat{{\bf Y}}]\varepsilon^{k} - ({\bf y}_{n} - {\bf u}^{k}_{m'_k})
  \end{split}
\end{equation}
be residual vectors of the linear regressions problem in Eq. \eqref{eq:2}. Then the classification result is determined by the \textit{residuals} directly. Moreover, all virtual faces will eventually be transformed into the \textit{residual} space, with which the related distances $d_r$ and unrelated $d_u$ can be defined as

\begin{equation*}
\begin{split}
  d_r({\bf {\bf X}}_k, {\bf Y}) = \| [\hat{{\bf X}}_k, -\hat{{\bf Y}}]\gamma^{k} - ({\bf y}_{n} - {\bf x}^k_{m_k}) \| = \|{\bf e}_r^{k}\|, \\
  d_u({\bf {\bf X}}_k, {\bf Y}) = \| [\hat{{\bf U}}_k, -\hat{{\bf Y}}]\varepsilon^{k} - ({\bf y}_{n} - {\bf u}^{k}_{m'_k}) \| = \|{\bf e}_u^{k}\|.
\end{split}
\end{equation*}
The test set ${\bf Y}$ will be classified into the class with the minimum decision distance $d = d_r/d_u$. In other words, when we are classifying a test set ${\bf Y}$ with ground-truth label $l$, the related distance $d_r^{l}$ and the unrelated distance $d_u^{l}$ are expected to be relatively smaller and larger respectively.

However, the residuals shown above just define a classifier in PLRC, rather than the feature extractor. It has been widely agreed that discriminant features are usually embedded in lower-dimensional manifold subspace \cite{xu2017sliced}. In this perspective, we propose a novel approach that learns the discriminant information from both positive set pairs and negative set pairs. In fact, the regression problems in Eq. \eqref{eq:2} show that the residuals also represent the linearly independent parts of two sets. As the variations such as illuminations, postures are quite common in the image sets, the residual space mainly consists of the task-specific (e.g., the person-specific) discrepancy between the sets. Thus the discriminant criterion built on the residual space will be less disturbed by the noisy variations, and implemented effectively.

\subsection{Discriminant Residual Learning}\label{sec3.2}

Subspace learning methods generally aim to reduce the within-class scatter and enlarge between-class scatter simultaneously. In the image set recognition problems, these two scatters can be represented by the related and unrelated distances. Fig.~\ref{fig:flowchart} shows a rough outline of the proposed method. In Fig.~\ref{fig:flowchart}(a), the intersection of \textit{Hawking} and \textit{Anka} is not empty, and classifier may make the wrong decision in this case. DRA extracts the useful information we are interested in, which allows the negative set pairs (i.e., \textit{Anka} and \textit{Smith}) to push the \textit{Hawking} away and makes the intra-class samples more compact as it shown in Fig.~\ref{fig:flowchart}(b). Finally, there is no overlap between the groups formed by \textit{Hawking} and \textit{Anka} in Fig.~\ref{fig:flowchart}(c). Overall, how to define the two scatters and find such feature subspace is a fundamental issue.

To facilitate derivation of the DRA model, we define the \textit{distance of interest} (DOI) as follows.
\begin{definition}[DOI]\label{def1}
For a single test set with ground-truth label $l$, the distance $d^{l}$ is named DOI among the total $c$ decision distances $d^{k}~(k=1,2,\ldots,c)$. Correspondingly, ${\bf e}^l_r$ and ${\bf e}^l_u$ are named \textit{interested residuals}.
\end{definition}

 As shown in Fig.~\ref{fig:flowchart}, the related and unrelated groups are built by the $l$-$th$ class (\textit{Hawking}), thus, the residuals ${\bf e}_u$ and ${\bf e}_r$ are the \textit{interested residuals}. The length of ${\bf e}_u$ is the related distance $d^l_r$ (shown in the blue arrow), and the length of ${\bf e}_r$ is the unrelated distance $d^l_u$ (shown in the orange arrow). Thus, the ratio of these two distances, i.e., $d^l_r/d^l_u$, is the DOI.

To make a correct classification under the above setting, the decision distance of the $l$-${th}$ class $d^{l}$ should be the smallest among $c$ distances. So $d^{l}$ is the most desirable one in all $c$ distances. Principally, DRA attempts to satisfy the above condition as much as possible by learning the features of \textit{interested residuals}.

To extract discriminant features as much as possible, we propose to learn a discriminative and low-dimensional subspace ${\bf P}$ from the regression errors. In such a discriminant embedding subspace, distance between the related group and the test set ${\bf Y}$ can be written as $\|{\bf P}^T{\bf e}_r^{k}\|$, while distance between the unrelated group and the test set is $\|{\bf P}^T{\bf e}_u^{k}\|$. Now a Rayleigh quotient-like distance is proposed in the subspace as
\begin{equation}\label{eq:proj_DOI}
\tilde{d}^{k}=\frac{\| [\hat{{\bf X}}_k, -\hat{{\bf Y}}]\gamma^{k} - ({\bf y}_{n} - {\bf x}^k_{m_k}) \|}{\| [\hat{{\bf U}}_k, -\hat{{\bf Y}}]\varepsilon^{k} - ({\bf y}_{n} - {\bf u}^{k}_{m'_k}) \|}\triangleq
\frac{\| {\bf P}^T{\bf e}_r^{k}\|}{\| {\bf P}^T{\bf e}_u^{k}\|}.
\end{equation}
In this sense, the virtual faces are no longer needed, and our work focuses on the modeling of \textit{residual} representations.

The DRA method starts from the residual space and learns the features in low-dimensional discriminant subspace. Suppose there are $c$ test sets $({\bf Y}_1,{\bf Y}_2,\ldots,{\bf Y}_c)$ with ground-truth labels $(1,2,\dots,c)$ on hand. Let $d^{kl}$ be the distance between the training set ${\bf {\bf X}}_k$ and the test set ${\bf Y}_l$, while ${\bf e}^{kl}_{r}$ and ${\bf e}^{kl}_{u}$ be the regression residuals ($k,l=1,2,\ldots,c$). Then the distance in discriminant subspace can be represented as:
\begin{equation*}
\tilde{d}^{kl}=\frac{\parallel {\bf P}^T{\bf e}_r^{kl} \parallel}{\parallel {\bf P}^T{\bf e}_u^{kl} \parallel}.
\end{equation*}

Since the DOI is not always the minimum distance, our goal is to increase the likelihood of its occurrence. Intuitively the DOI $\tilde{d}^{ll}$ needs to be smaller than others in the discriminant subspace. As the distance is obtained from the difference between \textit{residual} representations, it can be taken as an error term. Accordingly, DRA proposes two discriminant models, i.e., the \textbf{Partial-Error} (PE) model and the \textbf{Total-Error} (TE) model.

In the PE model, we pay our attention only to the \textit{interested residuals} ${\bf e}^{ll}_{r}$ and ${\bf e}^{ll}_{u}$, and calculate the projection matrix ${\bf P}$ by solving the optimization problem as follows:
\begin{equation}\label{eq:4}
{\bf P}^* = \mathop{\arg\min}_{{\bf P}^T{\bf P}={\bf I}}\frac{\sum_{i=1}^c\parallel {\bf P}^T{\bf e}_r^{ii} \parallel^2}{\sum_{i=1}^c\parallel {\bf P}^T{\bf e}_u^{ii} \parallel^2}.
\end{equation}
Further, if we equip Eq.~\eqref{eq:4} with the Euclidean norm then it is equivalent to the following trace-ratio optimization problem:
\begin{equation*}
{\bf P}^* = \mathop{\arg\min}_{{\bf P}^T{\bf P}={\bf I}}\frac{\sum_{i=1}^c tr({\bf P}^T{\bf e}_r^{ii}{{\bf e}_r^{ii}}^T{\bf P})}{\sum_{i=1}^c tr({\bf P}^T{\bf e}_u^{ii}{{\bf e}_u^{ii}}^T{\bf P})}.
\end{equation*}

\begin{figure*}[htb]
\begin{center}
\includegraphics[width=0.9\linewidth]{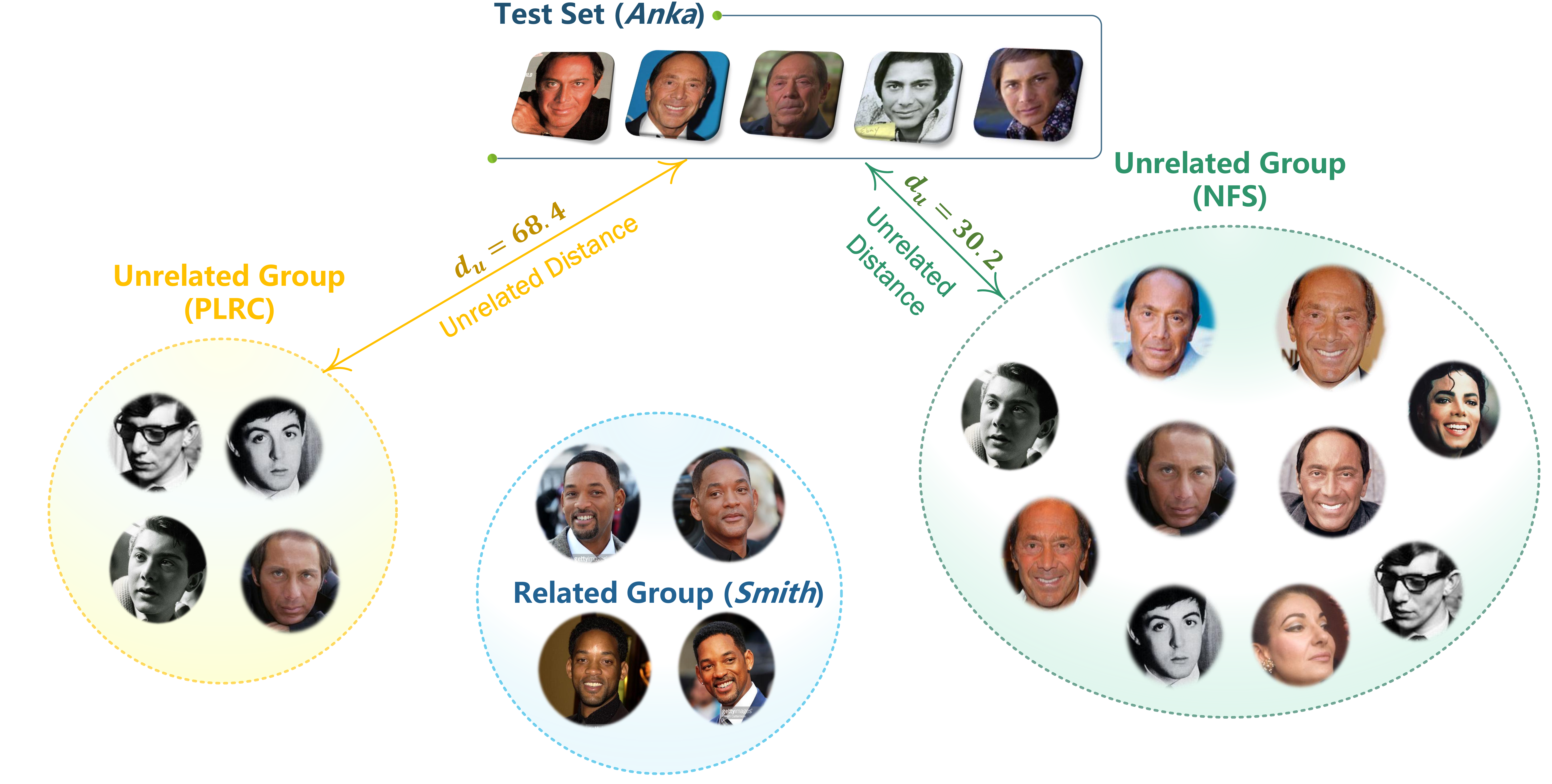}
\end{center}
\vspace{-1pc}
\caption{Unrelated groups generated by NFS and PLRC when the test set label (\textit{Anka}) is different to the training set label (\textit{Smith}). Ideally, all \textit{Anka} images should be grouped into the unrelated group to produce a smaller unrelated distance. In PLRC method, some \textit{Anka} images may not be properly grouped into the unrelated group due to style, color or other possible reasons. Consequently, the unrelated distance derived from PLRC is larger than that from NFS.}
\label{fig:NFS_Illustration}
\end{figure*}

It can be approximated by a generalized eigenvalue decomposition (GEVD) problem, i.e.,
\begin{equation}\label{eq:5}
\left( \sum_{i=1}^c{\bf e}_u^{ii}{{\bf e}_u^{ii}}^T \right) {\bf p}=\lambda \left( \sum_{i=1}^c{\bf e}_r^{ii}{{\bf e}_r^{ii}}^T \right) {\bf p}.
\end{equation}
The eigenvectors corresponding to the first $t$ largest eigenvalues are the so-called dominant eigenvectors which constitute the projection matrix ${\bf P}=[{\bf p}_1,{\bf p}_2,\ldots,{\bf p}_t]\in{\mathbb R}^{d \times t}$.

The TE model takes all residual vectors into account. In the best-case scenario, the DOIs, i.e., $\tilde{d}^{ll}$, are expected to be smaller than distances $\tilde{d}^{kl}~(k\ne l)$. Equivalently, $1/\tilde{d}^{kl}~(k\ne l)$ should be smaller than $1/\tilde{d}^{ll}$ in the subspace. We call the union of distances $1/\tilde{d}^{kl}~(k\ne l)$ and DOIs the total distance set. Now all distances in the total distance set are required to be smaller under the discriminant projection. Then the objective function can be written as
\begin{equation*}
{\bf P}^* = \mathop{\arg\min}_{{\bf P}^T{\bf P}={\bf I}}\frac{\sum_{i=1}^c \parallel {\bf P}^T{\bf e}_r^{ii} \parallel^2 + \sum_{i=1}^c\sum_{j\ne i} \parallel {\bf P}^T{\bf e}_u^{ij} \parallel^2}
{\sum_{i=1}^c \parallel {\bf P}^T{\bf e}_u^{ii} \parallel^2 + \sum_{i=1}^c\sum_{j\ne i} \parallel {\bf P}^T{\bf e}_r^{ij} \parallel^2}.
\end{equation*}
Analogously, the TE model can also be converted to a GEVD problem as:
\begin{equation}\label{eq:6}
{\bf A}_1{\bf p}=\lambda{\bf A}_2{\bf p},
\end{equation}
where
\begin{equation} \label{eq:7}
    \begin{split}
    {\bf A}_1 \triangleq \sum_{i=1}^c{\bf e}_u^{ii}{{\bf e}_u^{ii}}^T + \sum_{i=1}^c\sum_{j\ne i}{\bf e}_r^{ij}{{\bf e}_r^{ij}}^T , \\
    {\bf A}_2 \triangleq \sum_{i=1}^c{\bf e}_r^{ii}{{\bf e}_r^{ii}}^T + \sum_{i=1}^c\sum_{j\ne i}{\bf e}_u^{ij}{{\bf e}_u^{ij}}^T .
    \end{split}
\end{equation}
The $t$ dominant eigenvectors are selected to form the projection matrix ${\bf P}$. Note that the lengths of interested residuals ${\bf{e}}_u^{ii}$ are expected to be larger than those of ${\bf{e}}_u^{ij}$ $(j\neq i)$, and the lengths of ${\bf{e}}_r^{ii}$ be smaller than those of ${\bf{e}}_r^{ij}$ $(j\neq i)$. Therefore, $\textbf{A}_1$ contains the distances that need to be maximized, i.e., the DOIs $d_u^{ii}$ and $d_r^{ij} (j\neq i)$. Similarly, $\textbf{A}_2$ consists of the distances to be minimized, i.e., the DOIs $d_r^{ii}$ and $d_u^{ij} (j\neq i)$.

With the help of discriminant projections, DRA makes its efforts to reduce the related distance while enlarge the unrelated distance. It should be noted that the ground-truth label of the test set is unknown in real scenarios. So the validation set is used to complete the learning process.

\subsection{A Nonfeasance Strategy to Build Unrelated Groups}\label{sec3.3}

 We now propose NFS, which provides another approach to construct the unrelated groups. Recall that given a test set with ground-truth label $l$ and a training set ${\bf X} = [{\bf {\bf X}}_1, {\bf {\bf X}}_2, \cdots, {\bf {\bf X}}_c]$ with $c$ classes, PLRC first chooses $m_k$ samples being closest to the test set based on their distance metric in~\cite{PLRC}. Such a sample selection approach sometimes makes errors.

 Specifically, within the $c$ unrelated distances $d_u^{1},d_u^{2},\ldots,d_u^{c}$ mentioned in Section \ref{sec3.1}, the unrelated DOI $d_u^{l}$ is expected to be the largest, while distances $d_u^{k}~(k\ne l)$ are expected to be smaller than $d_u^{l}$. Since ${\bf {\bf X}}_l$ is the closest set to ${\bf Y}_l$, the unrelated groups ${\bf {\bf U}}_{k}~(k\ne l)$ should contain as many samples of ${\bf {\bf X}}_l$ as possible to obtain a smaller $d_u^{k}~(k\ne l)$. Unfortunately, ${\bf {\bf X}}_l$ cannot guarantee its distance being always the smallest one. It indicates that there exists at least one unrelated subspaces ${\bf {\bf U}}_{k}~(k\ne l)$ containing few samples of ${\bf {\bf X}}_k$, which makes the unrelated distance $d_u^{k}~(k\ne l)$ large.

Now we present a synthetic exemplar. Let ${\bf x}_1$, ${\bf x}_2$, ${\bf x}_3$ and ${\bf y}$ be four samples in $\mathbb{R}^3$, and their ground-truth labels are 1, 2, 3, and 2, respectively.
\begin{equation*}
\begin{array}{cccc}
{\bf x}_1 & {\bf x}_2 & {\bf x}_3 & {\bf y}
\\
\begin{bmatrix}   1 \\ 0 \\ 0 \end{bmatrix} &
\begin{bmatrix}   2 \\ 2 \\ 2 \end{bmatrix} &
\begin{bmatrix}   0 \\ 1 \\ 0 \end{bmatrix} &
\begin{bmatrix}   1 \\ 1 \\ 1 \end{bmatrix}
\end{array}
\end{equation*}
Under the Euclidean distance, when we are constructing the unrelated group of ${\bf x}_3$, sample ${\bf x}_1$ will be probably chosen rather than ${\bf x}_2$, since
 \begin{equation*}
   \| {\bf x}_1 - {\bf x}_3 \|_2 = \sqrt{2} < \| {\bf x}_2 - {\bf x}_3 \|_2 = 3.
\end{equation*}
But ${\bf x}_2$ is actually better because the true label of ${\bf y}$ is 2. From another perspective, choosing ${\bf x}_1$ or ${\bf x}_3$, or even both, is not important when constructing the unrelated group of ${\bf x}_2$, since the regression residuals is almost the same. Moreover, it is more complicated in higher dimensional space. 

In this case, we can avoid such mistakes by removing the selection step. Given ${\bf X}=[{\bf {\bf X}}_1,{\bf {\bf X}}_2,\cdots,{\bf {\bf X}}_c]$, the related and unrelated subspaces of $k$-$th$ class are spanned by ${\bf {\bf X}}_k\in{\mathbb R}^{d \times m_k}$ and ${\bf {\bf U}}_{k}=[{\bf {\bf X}}_1,\ldots,{\bf {\bf X}}_{k-1},{\bf {\bf X}}_{k+1},\ldots,{\bf {\bf X}}_{c}]\in{\mathbb R}^{d \times (m-m_k)}$, respectively. Note that ${\bf U}_{k}$ is independent to the test set ${\bf Y}_l$, so the step of calculating distance in PLRC is not required in NFS.

Fig.~\ref{fig:NFS_Illustration} shows the unrelated groups generated by NFS and PLRC. Because of their style, color or other complicated factors, PLRC fails to take all \textit{Anka} images into the unrelated group. Consequently, the \textit{McCartney} and \textit{Hawking} images are considered to be members of an unrelated group, resulting in a larger unrelated distance. However, NFS avoids this scenario by using all the remaining samples. Thus, the unrelated distance $d_u^{k}$ $(k\ne l)$ of NFS will be smaller than that of PLRC; precisely, $d_u^{NFS} = 30.2 < 68.4 = d_u^{PLRC}$. As a result, the distance $d^{k}$ $(k\ne l)$ of NFS will be larger than that of PLRC.

In practice, NFS does not bring too much change to the DOI $d_u^l$, but it helps $d_u^{k}$ $(k\ne l)$ get smaller. After defining the unrelated group, the next procedure is to obtain the distances by solving the regression problems in Eq. \eqref{eq:2}.

Note that NFS is a new approach in constructing the unrelated groups in the proposed DRA framework. The ablation study with respect to real performance, between NFS and other related methods such as DLRC (only related groups) and PLRC (both related and unrelated groups), will be shown in the section of experiments.

\subsection{Regularization}\label{sec3.4}

In the high-dimensional space, the generalized eigenvalue problems shown in  Eq.~\eqref{eq:5} and  Eq.~\eqref{eq:6} may be singular. We introduce two regularization methods to address it in this section. To get a more general solution, the GEVD problem is written as
\begin{equation}\label{eq:8}
{\bf A}_1{\bf p} =\lambda {\bf A}_2{\bf p},
\end{equation}
where ${\bf A}_1, {\bf A}_2\in{\mathbb R}^{d\times d}$ are symmetrical and positive semi-definite.

We can add $\mu {\bf I}$ to matrix ${\bf A}_2$, where ${\bf I}$ is the identity matrix and $\mu>0$ is the regularization parameter. Then Eq.~\eqref{eq:8} can be written as a symmetric eigenvalue problem:
\begin{equation}\label{eq:9}
{\bf A}_1{\bf p} =\tilde{\lambda} ({\bf A}_2+\mu {\bf I}) {\bf p},
\end{equation}
where $\tilde{\lambda}={\bf p}^T {\bf A}_1 {\bf p}/({\bf p}^T {\bf A}_2 {\bf p} + \mu)$ is the perturbed $\lambda$ and $\tilde{\lambda} = \lambda$ when the perturbation $\mu=0$.

Another regularization strategy is the matrix exponential transformation. Actually, there is a nice property that if ${\bf A}_1$ is symmetric, then ${\exp}({\bf A}_1)$ is positive definite. As a result, the regularized eigenvalue problem based on matrix exponential can be described as the following two equivalent forms:
\begin{gather}{}
{\rm exp}({\bf A}_1){\bf p} =e^{\lambda} {\rm exp}({\bf A}_2){\bf p}.\label{eq:10}
\end{gather}
Therefore, the GEVD problems Eq.~(\ref{eq:9}) and Eq.~(\ref{eq:10}) are no longer singular. For convenience, they are abbreviated by \textbf{eig} and \textbf{exp} in the following parts, respectively.

The main steps of the DRA algorithm with the TE model are summarized in Algorithm~\ref{alg:DRA-TE}. The PE model can be embedded into the algorithm in a similar manner.

\begin{algorithm}
\caption {DRA-TE for Image Set Classification}\label{alg:DRA-TE}
\label{alg1}
\begin{algorithmic}[1]
\REQUIRE {Training sets ${\bf X}=[{\bf {\bf X}}_1,\cdots,{\bf {\bf X}}_c]\in{\mathbb R}^{d \times m}$, Validation sets ${\bf Q}=[{\bf Q}_1,\ldots,{\bf Q}_c]\in{\mathbb R}^{d \times M}$, Test set ${\bf Y}\in{\mathbb R}^{d \times n}$, Projection dimension $t$;}
\ENSURE {Projection matrix ${\bf P}\in{\mathbb R}^{d \times t}$,  Predicted label $\hat{k}$;}\\
\% \textit{Training Stage}\\
\FOR {$k,l=1,\cdots,c$}
\STATE  Get the unrelated groups ${\bf U}_k$ based on $\{{\bf {\bf X}}_k, {\bf Q}_l\}$;
\STATE  Solve the regression problems in Eq.~\eqref{eq:2};
\STATE  Obtain residuals ${\bf e}_r^{kl}$ and ${\bf e}_u^{kl}$ via Eq.~\eqref{eq:3};
\ENDFOR
\STATE  Form the GEVD problem in Eqs.~\eqref{eq:6}-\eqref{eq:7};
\STATE  Eigenvalue regularization via Eq.~\eqref{eq:9} or Eq.~\eqref{eq:10};
\STATE  Calculate $t$ dominant eigenvectors ${\bf p}_1,{\bf p}_2,\cdots,{\bf p}_t$, then the projection matrix ${\bf P}=[{\bf p}_1,{\bf p}_2,\cdots,{\bf p}_t]$;\\
\% \textit{Testing Stage}\\
\FOR {$k=1,\cdots,c$}
\STATE  Get the unrelated groups ${\bf U}_k$ based on $\{{\bf {\bf X}}_k, {\bf Y}\}$;
\STATE  Solve the regression problems in Eq.~\eqref{eq:2};
\STATE  Obtain ${\bf P}^T{\bf e}^k_r$, ${\bf P}^T{\bf e}^k_u$, and $\tilde{d}^k$ via Eq.~\eqref{eq:proj_DOI};
\ENDFOR
\STATE  Return the prediction $\hat{k} = \arg\underset{k}\min\:\: \{\tilde{d}^k\}$;
\end{algorithmic}
\end{algorithm}

\subsection{Complexity Analysis}\label{sec3.5}
For convenience, we denote the size of training set, validation set and test set by $n_{tr}$, $n_{va}$ and $n_{te}$, respectively. In the training stage, the main computation burden concentrates on the regression problems (which can be solved in a parallel manner) and exponential eigenvalue problem of order $d$, which require about $\mathcal{O}(c(min\{d,n_{tr}+n_{va}\})^3)$ and $\mathcal{O}(d^3)$ flops \cite{golub2013matrix}, respectively. Actually, as reported by Wu et al. \cite{wu2017inexact}, the computation cost of exponential generalized eigenvalue problem can be reduced to $\mathcal{O}(d(n_{tr}+n_{va})^2)$ by applying the krylov subspace methods.

In the test stage, the complexity of DRA method is $\mathcal{O}(c(min\{d,n_{tr}+n_{te}\})^3)$ which comes from $c$ regression problems. For DARG \cite{wang2018discriminant}, it takes about $\mathcal{O}(dn_{tr}^2+n_{tr}^3)$ to compute the kernel matrices and eigenvalue problem during training; and the testing complexity is $\mathcal{O}(dn_{te}^2)$. For PDL \cite{wang2017prototype}, the training complexity is $\mathcal{O}(d^2 m_{p})$ for each iteration of the optimization, where $m_{p}$ is the total number of prototypes; and the testing complexity of NN classifier is $\mathcal{O}(d m_{p} n_{te})$.

\section{Experiments and Analysis}\label{sec4}

In this section, extensive experiments are conducted to evaluate the DRA method. Several state-of-the-art methods including DLRC \cite{DLRC}, PLRC-\uppercase\expandafter{\romannumeral1} \cite{PLRC}, PLRC-\uppercase\expandafter{\romannumeral2} \cite{PLRC},  AHISD \cite{HISD}, CHISD \cite{HISD}, PDL \cite{wang2017prototype}, DARG \cite{wang2018discriminant}, Regularized Hull based ISCRC (RH-ISCRC) \cite{RISCRC}, Kernelized Convex Hull based ISCRC (KCH-ISCRC) \cite{RISCRC} and Probabilistic CRC (ProCRC) \cite{ProCRC}, NAN \cite{yang2017neural}, IDLM \cite{shah2016iterative}, are used to compare with DRA. Note that PLRC-I and PLRC-II construct the unrelated groups according to the weighted distance based on Tikhonov regularization and the Euclidean distance, respectively.

\begin{figure}[htb]
\begin{center}
\includegraphics[width=0.95\linewidth]{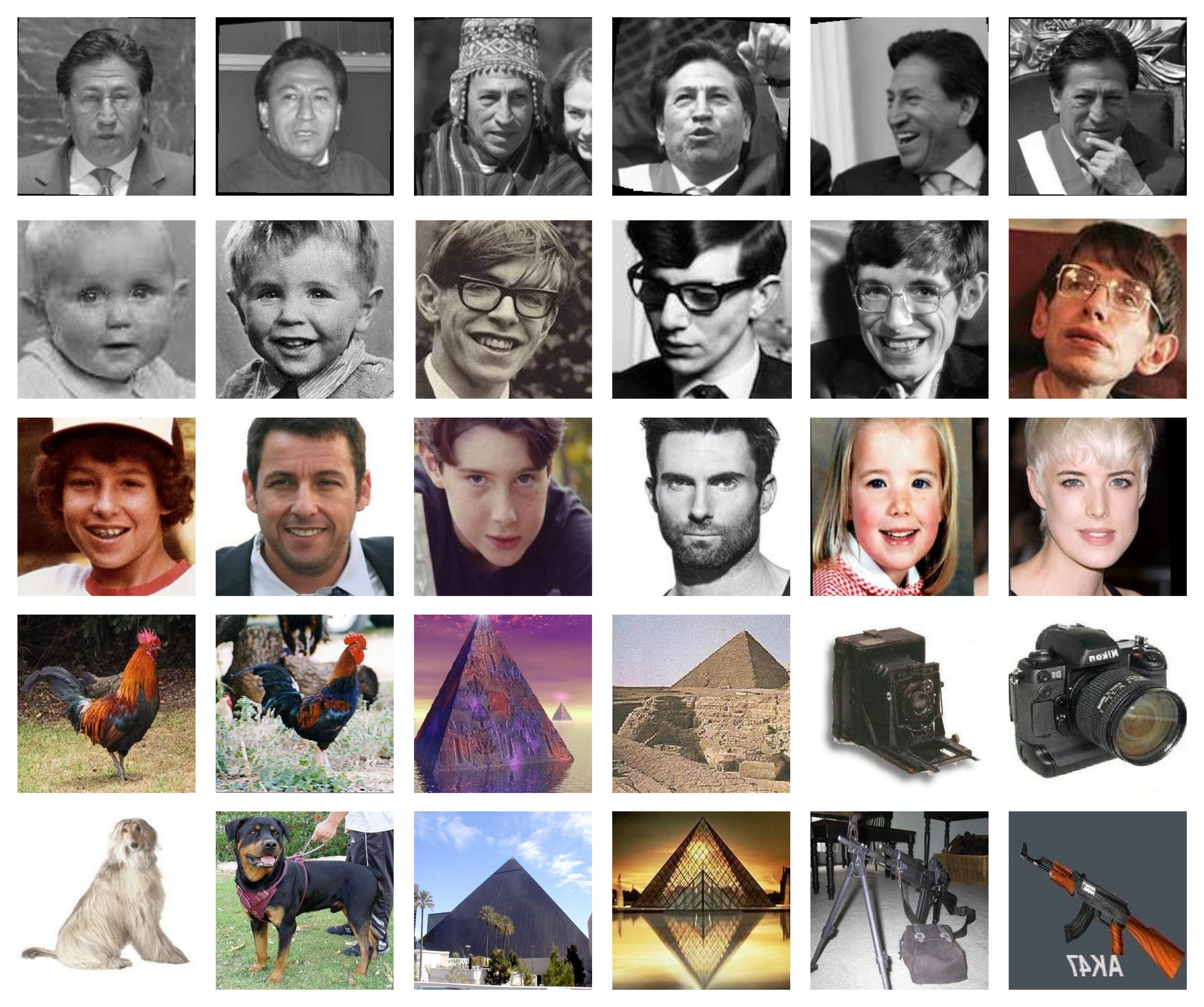}
\end{center}
\vspace{-1pc}
\caption{Illustrative images used in the experiments. From top to bottom: LFW, AgeDB, LAG, Caltech101 and Caltech256.}
\label{fig:illustrate-images}
\end{figure}

\subsection{Datasets and Experiment Settings}\label{sec4.1}

Four benchmark datasets are used in the experiments, and they are briefly included as follows. Some illustrative examples are shown in Fig. \ref{fig:illustrate-images}.
\begin{itemize}
\item {\bf LFW}~\cite{LFW_Sur}:
     The alignment version LFW-a \cite{LFWa} is used for evaluation here. Following the same setting in Ref. \cite{DLRC}, all images have been cropped into $90\times78$ by removing 88 pixels from top, 72 pixels from bottom, and 86 pixel margins from both left and right sides. Then the categories with more than 20 pictures are selected as the subset of LFW-a. Therefore, there are 3032 images of 62 individuals.
\item {\bf Large Age-Gap}: The Large Age-Gap (LAG) database \cite{LAG} is constructed with 3828 images of 1010 individuals. The people's ages range from 0 to 80 years old and images of teenagers show great difference in appearance, both of these two points make the age gap in LAG database inconceivably large. To build the image set classification task, two schemes are designed. Scheme 1 contains 572 images of 50 persons, each person has no less than 10 images. Scheme 2 contains 1665 images of 215 persons, each person has no less than 6 images.
\item {\bf AgeDB}: AgeDB \cite{agedb} is a widely used \textit{age in-the-wild} dataset which is proposed recently. All 16488 images of 567 celebrities are processed manually, thus the age labels are clean which makes AgeDB different from previous age database. Moreover, the images in AgeDB are totally \textit{in-the-wild} and are collected from uncontrolled, real-world conditions. Thus, it is challenging to recognize faces with such huge variation in age and conditions. Similarly, 73 individuals having more than 40 images are selected here. In order to better understand how age affects the recognition performance, two schemes are designed and will be detailed later.
\item {\bf Caltech101}: Caltech101 is a challenging objection recognition database contains over 9000 images for 102 categories \cite{cal101}. Each category contains about 31 to 800 images and represents an object or background, such as butterfly, camera, watch, google background, etc. The original image size of Caltech101 is $300\times200$.
\item {\bf Caltech256}: Caltech256 \cite{cal256} is the extension Caltech101. It consists of 30608 images from 256 object categories and clutter. Each category contains at least 80 images. Compared with Caltech101, there is no left-right alignment and artifact, which makes this task harder.
\end{itemize}

\begin{figure*}[htb]
\begin{center}
\includegraphics[width=0.49\linewidth]{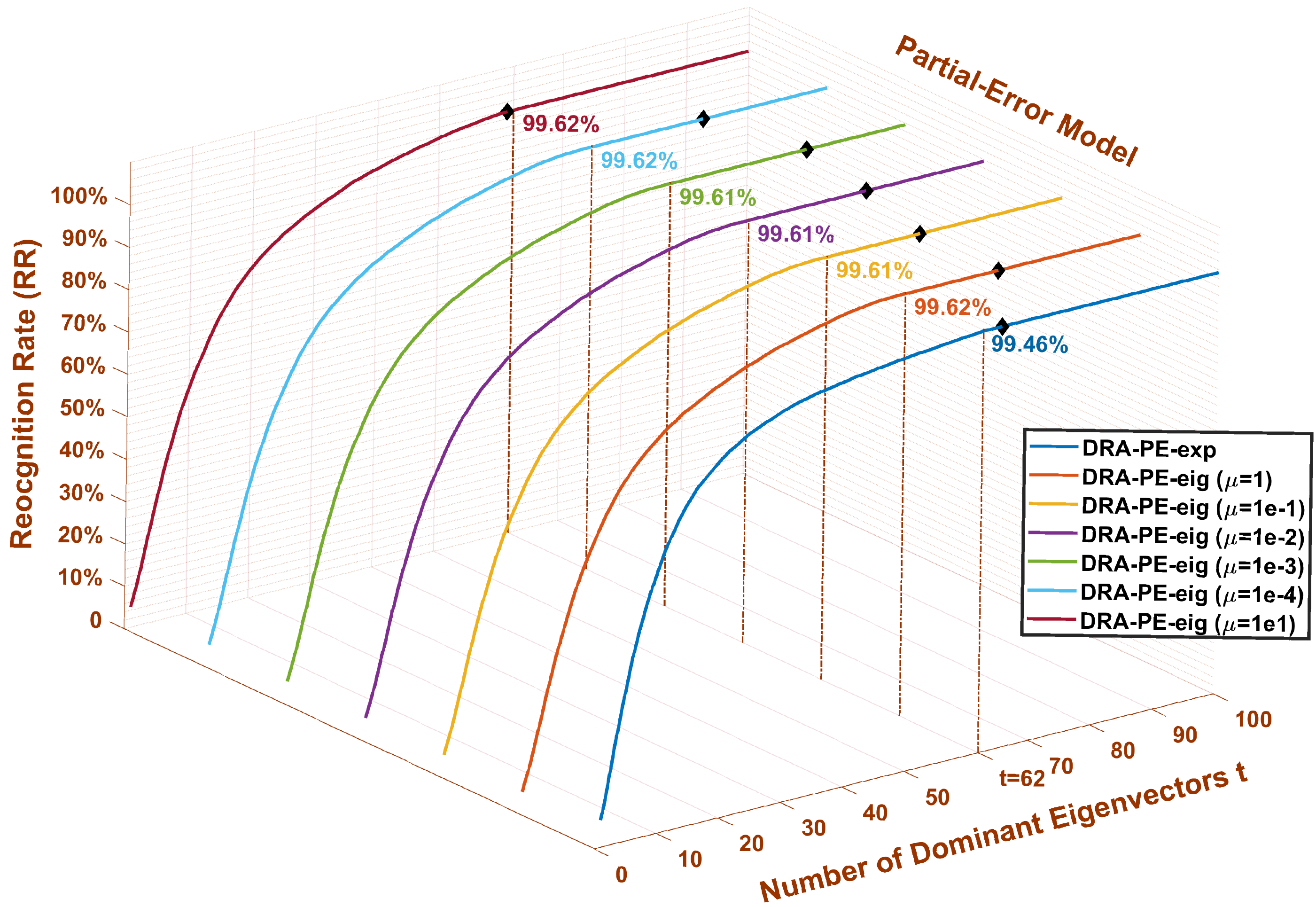}
\includegraphics[width=0.49\linewidth]{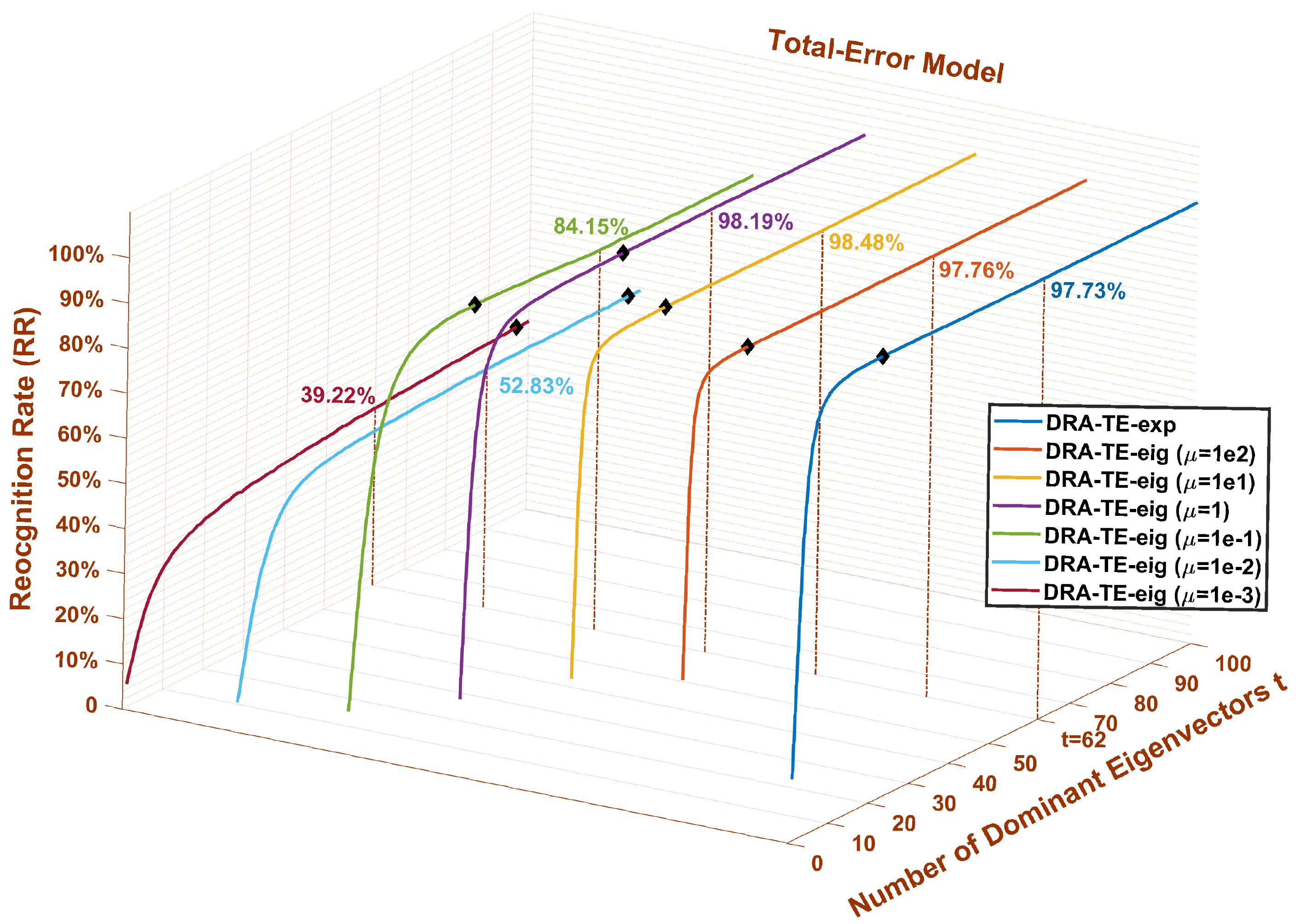}
\end{center}
\vspace{-1pc}
\caption{Recognition rate curves of the DRA method on LFW database (Top: PE model, Bottom: TE model). The accuracies when $t=c=62$ are given on each curve. The black diamond ``$\blacklozenge$'' means the highest recognition rate reached for the first time. $t=c=62$ is a good threshold and all curves will tend to be stable or falling when $t>c$. Compared with the PE model, the TE model owns a higher starting point and rapid rising curve with a lower upper limit.}
\label{fig:RR_curve_1}
\end{figure*}

\begin{table}[t]
\caption{General information of the datasets used in experiments.}\label{tab:Dataset}
\vspace{-1pc}
\begin{center}
\renewcommand{\tabcolsep}{0.6pc} 
\renewcommand{\arraystretch}{1.25} 
\begin{tabular}{ccccc}
\hline
& Datasets &  Type   & Deep Features & \\
\hline
& LFW \cite{LFWa}	    	&	Face	  & VggFace2-ResNet-50 \cite{VGGface2} & \\
& LAG \cite{LAG}	        		&	Face	   & VggFace2-ResNet-50 \cite{VGGface2} &  \\
& AgeDB \cite{agedb}	    		&	Face	   & VggFace2-ResNet-50 \cite{VGGface2} & \\
& Caltech101 \cite{cal101}    	&	Object	   & SE-ResNeXt-50 \cite{hu2018squeeze} &  \\
& Caltech256 \cite{cal256} & Object & SE-ResNeXt-50 \cite{hu2018squeeze} &  \\
\hline
\end{tabular}
\end{center}
\end{table}

Two kinds of deep neural network (DNN) \cite{DNN}, i.e., Residual Networks \cite{ResNet} and Squeeze-and-Excitation Networks \cite{hu2018squeeze}, are employed to learn deep features of face and object respectively. Compared with traditional hand-crafted features, DNN learns the powerful deep feature automatically and maps the raw data to a space where features are more separable \cite{VGGface2,hu2018squeeze,DNN}. For these two DNNs, the deep features are extracted from the penultimate layer. The feature dimensions of both DNNs are 2048. Since the input size of both two DNNs are 224$\times$224, all images are resized to fit it. General information of the datasets and features are shown in Tab.~\ref{tab:Dataset}. The network architectures are briefly summarized here.
\begin{itemize}
\item {\bf VggFace2-ResNet-50}: To capture the facial feature, the VggFace2 pretrained version of ResNet-50 is selected here. This model has been trained on the large-scale face database with more than 3.3 million images called \href{http://www.robots.ox.ac.uk/~vgg/data/vgg_face2/}{VggFace2}\footnote{\url{http://www.robots.ox.ac.uk/~vgg/data/vgg_face2/}} \cite{VGGface2}. 
\item {\bf SE-ResNeXt-50}: Squeeze-and-Excitation Networks is the champion of ILSVRC 2017 Image Classification Challenge \cite{hu2018squeeze}. One of the proposed templates named \href{https://github.com/hujie-frank/SENet}{SE-ResNeXt-50}\footnote{\url{https://github.com/hujie-frank/SENet}} are used for object feature extraction. 
\end{itemize}

\subsection{Hyper-parameter Selection}\label{sec4.2}

DRA requires the regularization parameter $\mu$ and the number of dominant eigenvectors $t$ in Algorithm~\ref{alg:DRA-TE}. In general, hyper-parameters are selected through cross-validation and empirical study.

\begin{figure*}[htb]
\begin{center}
\includegraphics[width=0.49\linewidth]{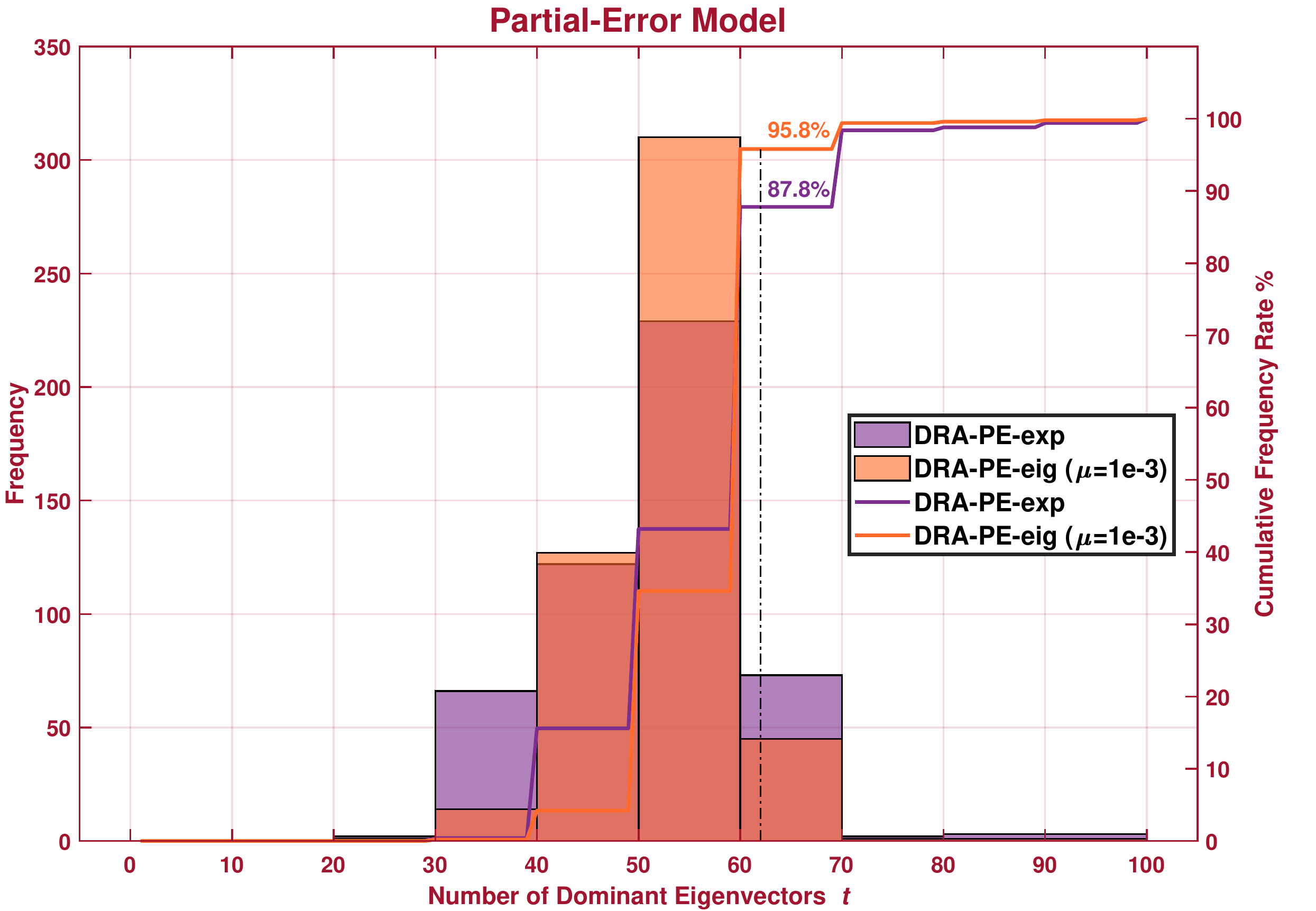}
\includegraphics[width=0.49\linewidth]{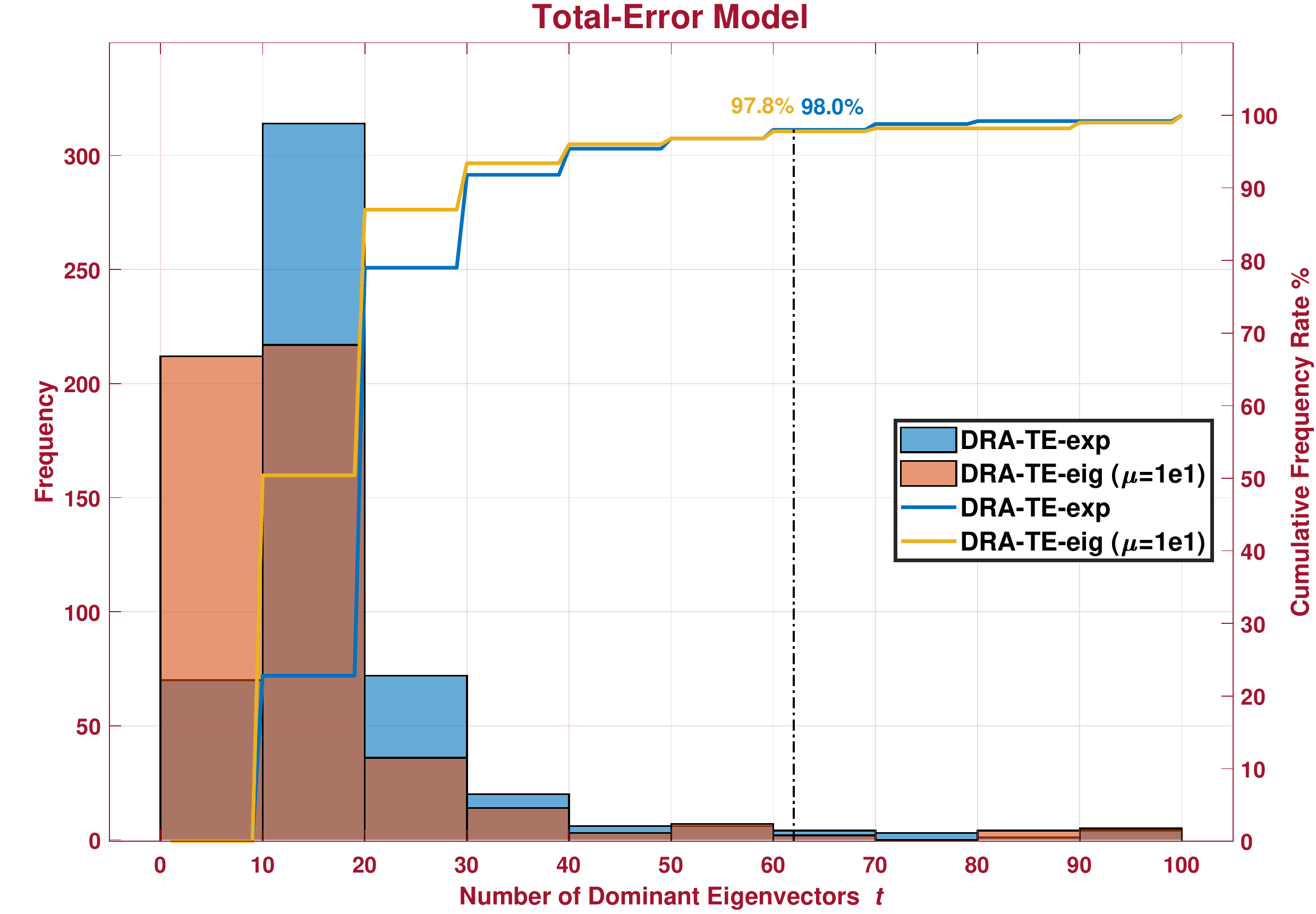}
\end{center}
\vspace{-1pc}
\caption{The frequency histogram and cumulative frequency rate curve of optimal $t$ value on the LFW database (Left: the PE model, Right: the TE model). The black dash line marks the cumulative frequency rate when $t=c=62$, and the value is also shown.}
\label{fig:HIST_1}
\end{figure*}

\begin{table}[htb]
\caption{Experiment results of 500 times random run on the LFW database.}\label{tab-LFW500run}
\vspace{-1pc}
\begin{center}
\renewcommand{\tabcolsep}{0.6pc} 
\renewcommand{\arraystretch}{1.25} 
\begin{tabular}{|c|c|c|c|c|c|}
\hline
\multicolumn{2}{|c|}{Methods} &\multirow{2}{*}{$\mu$} &\multicolumn{2}{c|}{Max Result} & $t=c$ \\
\cline{1-2} \cline{4-6}
Model & Regularization &  & t  & RR & RR  \\
\hline
\multirow{6}{*}{DRA-PE}  & exp & -    & 65 & 99.46\% & 99.46\% \\
                             & eig & 1e1  & 61 & 99.63\% & 99.62\% \\
                             & eig & 1    & 77 & 99.63\% & 99.62\% \\
                             & eig & 1e-1 & 77 & 99.64\% & 99.61\% \\
                             & eig & 1e-2 & 81 & 99.64\% & 99.61\% \\
                             & eig & 1e-3 & 84 & {\bf 99.64\%} & {\bf 99.62\%} \\
                             & eig & 1e-4 & 80 & 99.63\% & 99.62\% \\
\hline
\multirow{6}{*}{DRA-TE}      & exp & -    & 22 & 98.16\% & 97.73\% \\
                             & eig & 1e2  & 16 & 98.07\% & 97.75\% \\
                             & eig & 1e1  & 23 & {\bf 98.75\%} & {\bf 98.48\%} \\
                             & eig & 1    & 40 & 98.23\% & 98.19\% \\
                             & eig & 1e-1 & 31 & 85.82\% & 84.15\% \\
                             & eig & 1e-2 & 97 & 53.50\% & 52.83\% \\
                             & eig & 1e-3 & 97 & 41.68\% & 39.22\% \\
\hline
\end{tabular}
\end{center}
\end{table}

A series of experiments are conducted on the LFW-a database~\cite{LFWa}. The deep features acquired by VggFace2-ResNet-50 \cite{VGGface2} are used as the input. Two models of DRA with different regularizations are used here. The hyper-parameter $\mu$ is chosen from $\{1e1,1,1e-1,1e-2,1e-3,1e-4\}$ and $\{1e2,1e1,1,1e-1,1e-2,1e-3\}$ for the PE model and the TE model, respectively. The hyper-parameter $t$ is increased from 1 to 100 by 1. This experimental setting will be repeated randomly for 500 times.

Tab. \ref{tab-LFW500run} shows the mean recognition rate (RR) of 500 random experiments. In the Max Result column, the value of $t$ with the highest mean accuracy is shown for each setting. It turns out that $\mu=1e-3$ is the best value for the PE model and $1e1$ for the TE model. In addition, the recognition rates at $t=c=62$ are shown for comparison. The recognition rates of $t=c$ are only about 0.02\% and 0.27\% lower than the maximum of the PE model ($\mu=1e-3$) and the TE model ($\mu=1e1$), respectively. It means $t=c$ is a good approximation to the optimal $t$ value for both models.

Fig.~\ref{fig:RR_curve_1} shows the relationship between the number of dominant eigenvectors $t$ and the recognition rate (mean value of 500 experiments). At the beginning, the performance of both methods becomes better with the increase of $t$. As the recognition rate increases rapidly, all curves reach the maximum soon and then enter a stable phase. Therefore, all methods keep very little fluctuation in recognition rate and are insensitive to the changes of $t$ at this region. However, the recognition rate may decrease when $t$ exceeds a certain threshold, e.g., DRA-reg ($\mu=1e-1$) in the TE model. It indicates that $t=c$ is the best choice. According to the above results, the values of $\mu$ are selected as $1e-3$ and $1e1$ for the PE model and the TE model, respectively.

The frequency histogram and cumulative frequency curve of the optimal $t$ values are shown in Fig.~\ref{fig:HIST_1}. For the PE model, the frequency always reaches the peak at $t\!=\!c\!=\!62$. It indicates that $t\!=\!c$ is really suitable for the PE model. In general, distribution of $t$ in the TE model concentrates near a smaller value, which is consistent with the results shown in Fig.~\ref{fig:RR_curve_1}.

\begin{table*}[htb]
\caption{Recognition Rate (RR) and Standard Error (STE) on the LFW-a Dataset.}\label{tab_LFW}
\vspace{-1pc}
\begin{center}
\renewcommand{\tabcolsep}{1.2pc} 
\renewcommand{\arraystretch}{1.3} 
\begin{tabular}{|c|c|c|c|c|}
\hline
\multirow{2}{*}{Methods} &\multicolumn{3}{c|}{Raw Grey-Scale Images } & Deep \\
\cline{2-4}
                        & $10\times10$  & $15\times10$ & $30\times15$ & Features \\
\hline
AHISD        \cite{HISD}                                    & 39.78$\pm$1.06\% & 42.15$\pm$1.13\% & 44.57$\pm$0.99\%  & 95.59$\pm$0.47\%     \\
CHISD        \cite{HISD}                                    & 39.52$\pm$1.06\% & 41.94$\pm$1.13\% & 44.19$\pm$0.99\%  & 95.59$\pm$0.47\%    \\
DLRC         \cite{DLRC}                                    & 39.30$\pm$0.99\% & 42.04$\pm$1.13\% & 44.25$\pm$0.97\%  & 97.26$\pm$0.37\%    \\
RH-ISCRC     \cite{RISCRC}                                  & 61.77$\pm$1.06\% & 66.18$\pm$1.04\% & 68.66$\pm$1.04\%  & 98.12$\pm$0.37\%    \\
KCH-ISCRC    \cite{RISCRC}                                  & 44.09$\pm$1.01\% & 37.80$\pm$1.22\% & 56.02$\pm$1.22\%  & 97.42$\pm$0.49\%   \\
PLRC-\uppercase\expandafter{\romannumeral1}\cite{PLRC}      & 42.37$\pm$1.06\% & 46.08$\pm$1.08\% & 48.06$\pm$0.91\%  & 96.67$\pm$0.31\%   \\
PLRC-\uppercase\expandafter{\romannumeral2}\cite{PLRC}      & 38.98$\pm$0.93\% & 40.59$\pm$1.04\% & 42.85$\pm$1.01\%  & 96.94$\pm$0.38\%  \\
ProCRC        \cite{ProCRC}                                 & 46.45$\pm$1.13\% & 53.28$\pm$1.06\% & 55.38$\pm$1.04\%  & 89.89$\pm$0.69\%   \\
PDL        \cite{wang2017prototype}                         & 58.33$\pm$1.28\% & 63.23$\pm$1.13\% & 61.72$\pm$1.33\%  & 98.17$\pm$0.27\%   \\
DARG        \cite{wang2018discriminant}                     & 31.88$\pm$1.04\% & 35.00$\pm$1.13\% & 38.01$\pm$1.04\%  & 98.60$\pm$0.27\%   \\
NFS                                                         & 44.35$\pm$0.86\% & 48.71$\pm$0.97\% & 51.24$\pm$1.02\%  & 97.85$\pm$0.29\%   \\
\hline
ResNet-50 \cite{VGGface2}                                   & $-$ & $-$ & $-$ & 83.06$\pm$0.60\% \\
IDLM \cite{shah2016iterative}                               & $-$ & $-$ & $-$ & 91.77$\pm$0.64\% \\
NAN \cite{yang2017neural}                                   & $-$ & $-$ & $-$ & 93.39$\pm$0.66\% \\
\hline
DRA-PE-exp                                                  & 61.88$\pm$0.97\% & 67.53$\pm$1.06\% & 72.53$\pm$1.13\%  & 99.46$\pm$0.15\%    \\
DRA-TE-exp                                                  & 42.69$\pm$1.10\% & 49.14$\pm$0.93\% & 57.85$\pm$1.08\%  & 97.85$\pm$0.40\%    \\
DRA-PE-eig                                                  & {\bf 63.39$\pm$1.10\%} & {\bf 69.03$\pm$1.04\%} & {\bf 73.76$\pm$1.02\%}  & {\bf 99.73$\pm$0.11\%}   \\
DRA-TE-eig                                                  & 55.00$\pm$1.13\% & 63.76$\pm$1.13\% & 71.29$\pm$1.01\%  & 98.60$\pm$0.30\%   \\
\hline
\end{tabular}
\end{center}
\end{table*}

We observe that the optimal $t$ values tends to be smaller for the TE model, and more than 98\% of them are not larger than $c$. In Fig.~\ref{fig:RR_curve_1}, recognition rate of the TE model becomes stable as it reaches the maximum. Thus $t=c$ is usually in the stable phase and it is also a good alternative to the optimal setting.

Under the guidance of the above analysis, some hyper-parameter details in the following experiments are presented here. The two regularization methods shown in Eq.~\eqref{eq:9} and Eq.~\eqref{eq:10} are called DRA-eig and DRA-exp, respectively. NFS is used to construct the unrelated groups in DRA. Some parameters are set as below and unchanged throughout all experiments. The regression problems in Eq.~\eqref{eq:2} are solved by ridge regression, and the ridge parameter is empirically set to $1e-2$. The regularization parameter $\mu$ of DRA-reg in Eq.~\eqref{eq:9} is set as $1e-3$ for the PE model and $1e1$ for the TE model. For RH-ISCRC, the regularization parameters $\lambda_1 = \lambda_2 = 1e-3$. For AHISD and CHISD, the kernel version is selected since kernel mapping functions usually have better classification performance than the linear version. For ProCRC, we set $\gamma=1e-3$ and $\lambda=1e-2$, and equip it with vote mechanism to classify the set. Specifically, for each test image set, the prediction of this set is the class that achieves the maximum number of votes, where each vote is made by ProCRC based on single image in the set. The Gaussian kernel is uniformly used in above the kernel-based methods. For DARG, the kernel version based on Mahalanobis distance and Log-Euclidean distance is selected; the fusing coefficients are determined by cross-validation. For PLD, we follow the default hyper-parameters in its source code. For making a fair comparison, we replace the backbones of deep models (i.e., the IDLM and NAN methods) with the corresponding DNNs in our experiments. A baseline model is also designed. It builds three fully connected layers on the DNNs with a softmax classifier, where the final decision is made by voting.

\subsection{Performance on LFW-a Databases}\label{sec4.3}

In this section, comparison between the proposed methods and other state-of-the-art methods is presented, and the difference between two types of regularization in DRA is investigated. Furthermore, the performance on the raw images and the deep feature are evaluated. To this end, the raw grey-scale images of the LFW-a dataset is resized into $10\times10$, $15\times10$ and $30\times15$. All images are randomly divided into three parts: training set, validation set and test set. The number parameter previously mentioned is set as $(n_{train}, n_{valid}, n_{test})=(3,3,3)$, and the experiments are conducted randomly for 30 times. The average recognition rate (RR) and standard error (STE) are reported in Tab. \ref{tab_LFW}.

As shown in Tab. \ref{tab_LFW}, DRA-based methods are the best in recognition and more robust than other methods on the LFW-a dataset. The PE model outperforms the TE model by about 3\%-11\% in recognition accuracy. This performance difference may be attributed to their work principles. Specifically, the TE model not only strengthens the DOIs, but also penalizes the distances that not belong to DOIs. It means that the TE model involves more information with additional constrains, which may be too strong to be satisfied in some cases.

Though deep methods like IDLM and NAN learn a well representations for the set samples, their classifiers are simple vote rule and softmax regression. This leads to the moderate performance of 91.77\% and 93.39\% for IDLM and NAN, respectively.

\begin{figure}[t]
\begin{center}
\includegraphics[width=1.0\linewidth]{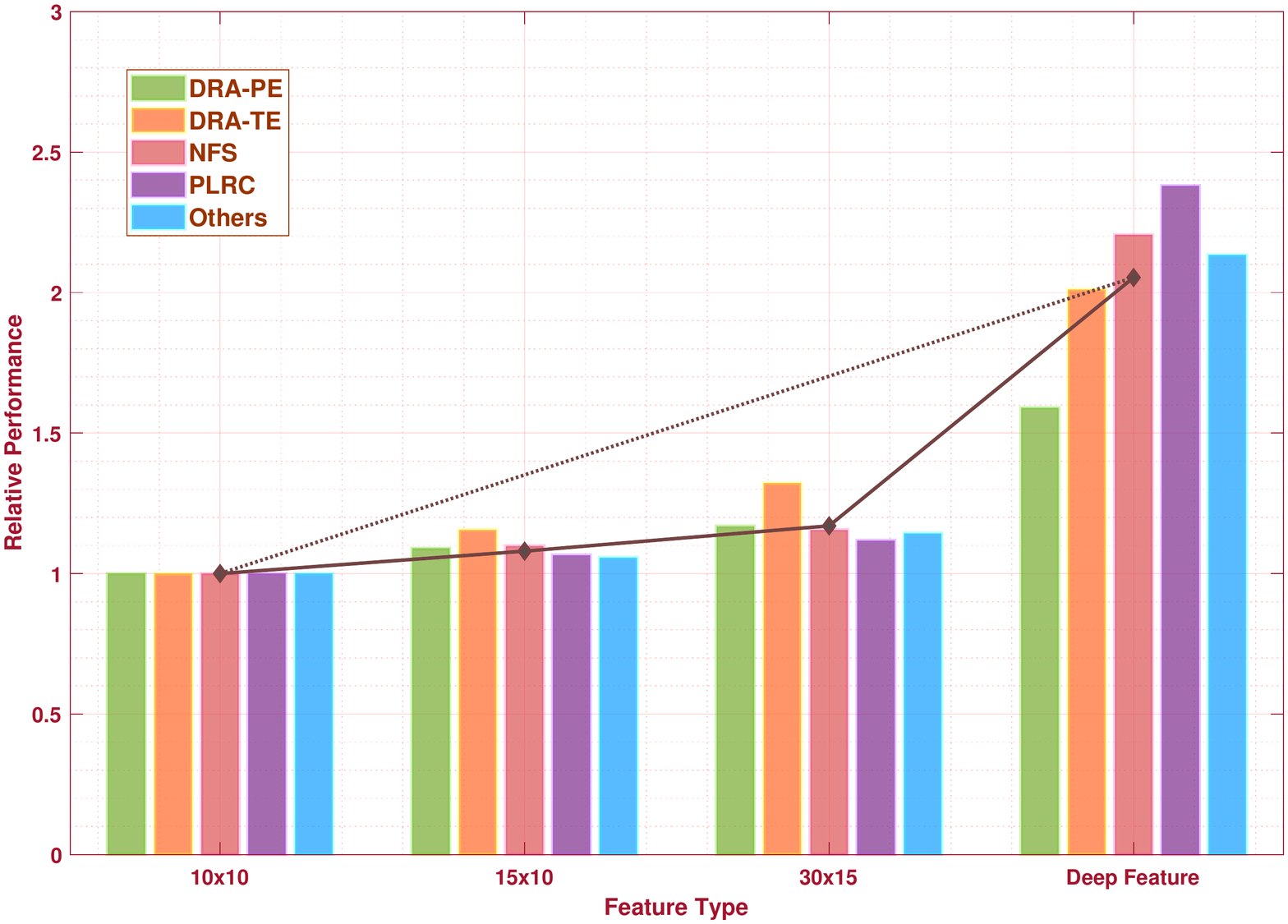}
\end{center}
\vspace{-1pc}
\caption{Comparison of relative performance between deep features and different pixels on the LFW-a dataset. The relative performance is computed as the ratio of current accuracy to baseline, which is obtained by using the 10$\times$10 image size. The \textbf{Others} (i.e., the blue bar) means the average performance of the compared methods. Though the deep features improve the accuracy significantly, the performance of our methods will not fluctuate drastically when feature change.}
\label{fig:improve_lfw}
\end{figure}

Another interesting result is that the deep features not only improve the classification performance, but also narrows the performance gap between different methods by making the feature space more separable. Precisely, the recognition rates on deep features are 8\%-20\% (even 60\% for DARG) higher than the maximum recognition rates on the raw image data. Fig.~\ref{fig:improve_lfw} shows the significant recognition improvement by deep features. Though the ascent of resolution improves the accuracy, such improvement is still too \textit{flat} compared with the results obtained by using deep features. Besides, as deep features are generally more discriminative and robust to noisy variations than raw pixels, and the accuracies of all methods are close to 100\% on deep features, the relative improvements will be mainly determined by the denominator. Thus, the relative improvement looks inversely proportional to the results of the raw pixels.

\subsection{Performance on LAG Database}\label{sec4.4}

As stated in Section \ref{sec4.1}, for LAG database, the average number of images per individuals in Scheme 2 is less than that in Scheme 1. Therefore, Scheme 2 can be considered as the \textit{small sample size} scenario to some extent. Under this setting, the number parameters are set as $(n_{train}, n_{valid}, n_{test})=(3,3,3)$ for Scheme 1 and $(2,2,2)$ for Scheme 2. Similarly, the experiments are random carried out for 30 times.

\begin{table}[htb]
\caption{Recognition Rate (RR) and Standard Error (STE) on LAG Database.}\label{tab_LAG}
\vspace{-1pc}
\begin{center}
\renewcommand{\tabcolsep}{1.0pc} 
\renewcommand{\arraystretch}{1.3} 
\begin{tabular}{|c|c|c|}
\hline
\multirow{2}{*}{Methods} &  \multicolumn{2}{c|}{Deep Features}\\
\cline{2-3}
                        &  Scheme 1 & Scheme 2 \\
\hline
AHISD        \cite{HISD}                                    & 83.33$\pm$1.84\% & 70.53$\pm$2.92\%   \\
CHISD        \cite{HISD}                                    & 83.33$\pm$1.84\% & 70.53$\pm$2.92\%  \\
DLRC         \cite{DLRC}                                    & 92.53$\pm$1.13\%  & 73.36$\pm$2.68\%  \\
RH-ISCRC     \cite{RISCRC}                                  & 94.93$\pm$0.75\% & 77.57$\pm$2.23\%  \\
KCH-ISCRC    \cite{RISCRC}                                  & 94.93$\pm$0.66\%  & 73.92$\pm$2.28\% \\
PLRC-\uppercase\expandafter{\romannumeral1}\cite{PLRC}      & 93.00$\pm$1.00\%  & 73.95$\pm$2.63\% \\
PLRC-\uppercase\expandafter{\romannumeral2}\cite{PLRC}      & 92.27$\pm$1.24\%  & 72.93$\pm$2.70\%\\
ProCRC        \cite{ProCRC}                                 & 86.13$\pm$1.31\%  & 57.49$\pm$1.94\% \\
PDL        \cite{wang2017prototype}                         & {\bf 98.47$\pm$0.27\%} & $-$   \\
DARG        \cite{wang2018discriminant}                     & 97.67$\pm$0.38\% & $-$  \\
NFS                                                         & 93.60$\pm$1.10\%  & 74.33$\pm$2.67\% \\
\hline
ResNet-50 \cite{VGGface2}                                   & 84.13$\pm$1.44\% & 51.29$\pm$1.82\% \\
IDLM \cite{shah2016iterative}                               & 92.27$\pm$0.99\% & 72.62$\pm$2.40\% \\
NAN \cite{yang2017neural}                                   & 96.67$\pm$0.96\% & 75.53$\pm$2.14\% \\
\hline
DRA-PE-exp                                                 & 98.20$\pm$0.40\% & 82.99$\pm$0.99\%  \\
DRA-TE-exp                                                 & 95.53$\pm$0.68\%  & 76.31$\pm$2.17\%  \\
DRA-PE-eig                                                 & 98.27$\pm$0.33\% & {\bf 87.55$\pm$1.04\%} \\
DRA-TE-eig                                                 & 96.27$\pm$0.58\%  & 81.50$\pm$2.19\% \\
\hline
\end{tabular}
\end{center}
\end{table}

As shown in Tab. \ref{tab_LAG}, the DRA-reg-PE method outperforms other methods except for PDL in scheme 1. In scheme 2, PDL and DARG methods do not work with very few samples. Besides, the difference in recognition rates between DRA and other competitors is at least 10\% in Scheme 2, and the standard deviations of DRA are significantly lower than others. It also turns out that DRA adapts well to sample size variations.

From another perspective, the inadequate sample size result in a decline in the recognition performance and the recognition rates are down by 10.7\%-28.6\%. Fortunately, DRA still achieves the recognition rate of 87.55\% which is significantly higher than others. So the robustness of our new method is reflected not only by the changes of training sample size, but also the variance of classification accuracy.

\subsection{Performance on AgeDB Database}\label{sec4.5}

\textit{Age in-the-wild} is another difficult issue in face recognition. As we all known, human facial appearance can change dramatically with different ages. In this section, the impact of the age gap on recognition accuracy is shown quantitatively. The experiments consist of two schemes with different age compositions. The number parameter for data partition is set as $(n_{train}, n_{valid}, n_{test})=(3,3,3)$ for both schemes as follows.

\begin{itemize}
\item {\bf Mix-Up}: In this scheme, samples of all ages of each class are mixed together, which means there is no processing for the raw data. Thus, the age factor does not shown up and it can be considered as the traditional \textit{in-the-wild} classification task.
\item {\bf Age-Gap}: To build the age-gap, the face images are arranged in ascending order of age for each class. The first half of images are used for training and validating. The last quarter is drawn out to form the test set. Therefore, the unused quarter is the age-gap expected. From another perspective, the training and validation samples represent a relatively young ages of the person; while the test images are older.
\end{itemize}

\begin{table}[htb]
\caption{Recognition Rate (RR), Standard Error (STE) and CPU Time(s) of the Mix-Up scheme on AgeDB Database.}\label{tab_AgeDB_Mixup}
\vspace{-1pc}
\begin{center}
\renewcommand{\tabcolsep}{0.8pc} 
\renewcommand{\arraystretch}{1.3} 
\begin{tabular}{|c|c|c|c|}
\hline
\multicolumn{4}{|c|}{\textbf{The Mix-Up Scheme}} \\
\multicolumn{4}{|c|}{VggFace2-ResNet-50 Deep Features} \\
\hline
\multirow{2}{*}{Methods} & \multirow{2}{*}{RR $\pm$ STE} & \multicolumn{2}{c|}{Time (s)}  \\
\cline{3-4}
                        &  & Training & Test\\
\hline
AHISD        \cite{HISD}                                     & 91.83$\pm$1.15\% & $-$ & 0.07	   \\
CHISD        \cite{HISD}                                     & 91.83$\pm$1.15\% & $-$ & 0.07	   \\
DLRC         \cite{DLRC}                                     & 94.75$\pm$0.57\% & $-$ & 0.01	    \\
RH-ISCRC     \cite{RISCRC}                                   & 95.21$\pm$0.46\% & 0.01 & 0.09     \\
KCH-ISCRC    \cite{RISCRC}                                   & 94.75$\pm$0.47\% & 0.01 & 0.13     \\
PLRC-\uppercase\expandafter{\romannumeral1}\cite{PLRC}       & 95.07$\pm$0.55\% & $-$ & 0.11	    \\
PLRC-\uppercase\expandafter{\romannumeral2}\cite{PLRC}       & 94.38$\pm$0.60\% & $-$ & 0.11	    \\
ProCRC        \cite{ProCRC}                                  & 89.13$\pm$0.82\% & $-$ & 0.01     \\
PDL        \cite{wang2017prototype}                          & 96.58$\pm$0.27\% & 98.56 &  0.03 \\
DARG        \cite{wang2018discriminant}                      & {\bf 97.99$\pm$0.27\%} & 38.45 & 0.01 \\
NFS                                                          & 95.48$\pm$0.46\% & $-$ & 0.66     \\
\hline
ResNet-50 \cite{VGGface2}                                   & 85.62$\pm$0.95\% & 115.58 & 0.05 \\
IDLM \cite{shah2016iterative}                               & 91.74$\pm$0.72\% & 149.10 & 0.04 \\
NAN \cite{yang2017neural}                                   & 97.91$\pm$0.65\% & 102.35 & 0.04 \\
\hline
DRA-PE-exp                                                  & 97.58$\pm$0.40\% & 65.92 & 0.66 	   \\
DRA-TE-exp                                                  & 95.75$\pm$0.46\% & 118.10 & 0.66 	    \\
DRA-PE-eig                                                  & 97.95$\pm$0.33\% & 1.63 & 0.66 	 \\
DRA-TE-eig                                                  & 96.94$\pm$0.46\% & 49.98 & 0.66 	   \\
\hline
\end{tabular}
\end{center}
\end{table}

We show the results under the Mix-Up setting in Tab.~\ref{tab_AgeDB_Mixup}. We can see that the best recognition accuracy, i.e., 97.99\%, is obtained by DARG. The results of DRA-PE-exp and DRA-PE-eig are 97.58\% and 97.95\%, respectively. These are very close to the accuracy of DARG. However, the training time of DRA-PE-eig is just 1.6s, which is much smaller than that of DARG (38.5s). It means that our method exhibits superiority in the efficiency perspective.

\begin{table}[htb]
\caption{Recognition Rate (RR), Standard Error (STE) and CPU Time(s) of Age-Gap scheme on AgeDB Database.}\label{tab_AgeDB_AgeGap}
\vspace{-1pc}
\begin{center}
\renewcommand{\tabcolsep}{0.8pc} 
\renewcommand{\arraystretch}{1.3} 
\begin{tabular}{|c|c|c|c|}
\hline
\multicolumn{4}{|c|}{\textbf{The Age-Gap Scheme}} \\
\multicolumn{4}{|c|}{VggFace2-ResNet-50 Deep Features} \\
\hline
\multirow{2}{*}{Methods} & \multirow{2}{*}{RR $\pm$ STE} & \multicolumn{2}{c|}{Time (s)}  \\
\cline{3-4}
                        &  & Training & Test\\
\hline
AHISD        \cite{HISD}                                     & 66.53$\pm$0.88\% & $-$ & 0.07   \\
CHISD        \cite{HISD}                                     & 66.53$\pm$0.88\% & $-$ & 0.06  \\
DLRC         \cite{DLRC}                                     & 74.79$\pm$0.73\% & $-$ & 0.01   \\
RH-ISCRC     \cite{RISCRC}                                   & 76.44$\pm$0.55\% & 0.01 & 0.09   \\
KCH-ISCRC    \cite{RISCRC}                                   & 74.34$\pm$0.79\% & 0.01 & 0.13   \\
PLRC-\uppercase\expandafter{\romannumeral1}\cite{PLRC}       & 74.93$\pm$0.62\% & $-$ & 0.11  \\
PLRC-\uppercase\expandafter{\romannumeral2}\cite{PLRC}       & 74.84$\pm$0.75\% & $-$ & 0.11  \\
ProCRC        \cite{ProCRC}                                  & 67.81$\pm$1.00\% & $-$ & 0.01   \\
PDL        \cite{wang2017prototype}                          & 75.02$\pm$0.93\% & 91.30 &  0.03\\
DARG        \cite{wang2018discriminant}                      & 80.18$\pm$0.64\% & 38.08 & 0.01 \\
NFS                                                          & 76.58$\pm$0.64\% & $-$ & 0.66  \\
\hline
ResNet-50 \cite{VGGface2}                                   & 65.84$\pm$0.99\% & 112.68 & 0.05 \\
IDLM \cite{shah2016iterative}                               & 71.42$\pm$0.81\% & 153.79 & 0.03 \\
NAN \cite{yang2017neural}                                   & 75.94$\pm$0.71\% & 99.64 & 0.04 \\
\hline
DRA-PE-exp                                                  & 81.1$0\pm$0.71\% & 65.84 & 0.67  \\
DRA-TE-exp                                                  & 77.35$\pm$0.69\% & 118.00 & 0.67    \\
DRA-PE-eig                                                  & {\bf 82.37$\pm$0.68\%} & 1.63 & 0.67  \\
DRA-TE-eig                                                  & 79.09$\pm$0.66\% & 49.94 & 0.67  \\
\hline
\end{tabular}
\end{center}
\end{table}

Tab. \ref{tab_AgeDB_AgeGap} presents the results under the Age-Gap setting. It shows a simulation on \textit{age in-the-wild} problem. Compared with the results obtained in the Mix-Up case, the accuracies of all methods decline rapidly, and the standard deviations of them become larger. The accuracy gap between these two schemes is about 15\%-25\%. In particular, the recognition accuracy of DRA-PE-eig is 82.37\%, which is far better than the results of several state-of-the-art methods. Taking PDL and DARG for example, their accuracies are just 75\% and 80.18\% under the same experimental setup. Therefore, these results together reflect the effectiveness of the discriminant residual analysis on the age-varied face set recognition.

\begin{table*}[htb]
\caption{Recognition Rate (RR), Standard Error (STE) and CPU Time(s) on Caltech-101 Database.}\label{tab_Cal}
\vspace{-1pc}
\begin{center}
\renewcommand{\tabcolsep}{1.2pc} 
\renewcommand{\arraystretch}{1.3} 
\begin{tabular}{|c|c|c|c|c|c|c|}
\hline
\multirow{3}{*}{Methods} &\multicolumn{3}{c|}{\textbf{SPM} ($d=3000$)} & \multicolumn{3}{c|}{\textbf{SE-ResNeXt-50} ($d=2048$)} \\
\cline{2-7}
                        & \multirow{2}{*}{RR $\pm$ STE} & \multicolumn{2}{c|}{Time (s)} & \multirow{2}{*}{RR $\pm$ STE} & \multicolumn{2}{c|}{Time(s)} \\
\cline{3-4}
\cline{6-7}
                        &                             & Training & Test &  & Training & Test \\
\hline
AHISD        \cite{HISD}            & 70.92$\pm$0.55\% & $-$ & 0.14 	 & 85.23$\pm$0.44\% & $-$ & 5.70    \\
CHISD        \cite{HISD}            & 70.92$\pm$0.55\% & $-$ & 0.14 	 & 85.23$\pm$0.44\% & $-$ & 5.70    \\
DLRC         \cite{DLRC}            & 70.88$\pm$0.55\% & $-$ & 0.02 	 & 85.13$\pm$0.42\% & $-$ & 0.01    \\
RH-ISCRC     \cite{RISCRC}     & 69.80$\pm$0.58\% & 0.02 & 4.05 	 & 85.88$\pm$0.44\% & 0.01 & 0.43    \\
KCH-ISCRC    \cite{RISCRC}   & 74.08$\pm$1.50\% & 6.45 & 0.12 	 & 80.59$\pm$1.32\% & 4.28 & 0.09    \\
PLRC-\uppercase\expandafter{\romannumeral1}\cite{PLRC}        & 72.19$\pm$0.58\% & $-$ & 0.57 	 & 84.97$\pm$0.49\% & $-$ & 0.42    \\
PLRC-\uppercase\expandafter{\romannumeral2}\cite{PLRC}        & 70.42$\pm$0.68\% & $-$ & 0.54 	 & 84.35$\pm$0.58\% & $-$ & 0.40    \\
ProCRC        \cite{ProCRC}                                   & 57.09$\pm$0.69\% & $-$ & 0.05 	 & 69.61$\pm$0.66\% & $-$ & 0.03    \\
PDL        \cite{wang2017prototype}                           & 69.25$\pm$0.64\% & 9610.34 & 0.11 	 & 82.71$\pm$0.77\% & 1664.33 & 0.06    \\
DARG        \cite{wang2018discriminant}                       & 62.58$\pm$0.68\% & 67.89 & 0.01 	 & 88.59$\pm$0.49\% & 68.71 & 0.01    \\
NFS                                                           & 72.84$\pm$0.53\% & $-$ & 3.90 	 & 86.90$\pm$0.44\% & $-$ & 2.82    \\
\hline
SE-ResNeXt-50 \cite{hu2018squeeze}                          & $-$ &$-$ & $-$ & 74.38$\pm$0.72\% & 191.05 & 0.08 \\
IDLM \cite{shah2016iterative}                               & $-$ &$-$ & $-$ & 83.50$\pm$0.63\% & 199.99 & 0.06 \\
NAN \cite{yang2017neural}                                   & $-$ &$-$ & $-$ & 88.50$\pm$0.59\% & 143.14 & 0.06 \\
\hline
DRA-PE-exp                                                   & 72.39$\pm$1.66\% & 122.32 & 3.91 	 & 87.35$\pm$0.42\% & 72.00 & 2.87    \\
DRA-TE-exp                                                  & 74.80$\pm$0.66\% & 531.00 & 3.91 	 & 85.00$\pm$0.44\% & 384.98 & 2.87    \\
DRA-PE-eig                                                   & 76.67$\pm$1.50\% & 6.82 & 3.91 	 & {\bf 89.80$\pm$0.47\%} & 4.06 & 2.87  \\
DRA-TE-eig                                                   & {\bf 79.51$\pm$0.53\%} & 404.66 & 3.91 	 & 85.20$\pm$0.42\% & 298.07 & 2.87     \\
\hline
\end{tabular}
\end{center}
\end{table*}

\subsection{Performance on Caltech101 Database}\label{sec4.6}

Object recognition is one of the important tasks in computer vision, and it could be harder than face recognition since the features are multifarious and difficult to extract. In this section, the Spatial Pyramid Matching (SPM)~\cite{SPM} and the Squeeze-and-Excitation Networks (SE-Net) \cite{hu2018squeeze} are employed as the representations of hand-crafted features and deep features, respectively. The number parameter is set as $(n_{train}, n_{valid}, n_{test})=(5, 5, 5)$. Finally, the average results of 30 experiments are reported in Tab. \ref{tab_Cal}.

When the SPM features are used to evaluate these methods, we can see that the best recognition accuracy of those compared methods is 74.08\%, which is obtained by KCH-ISCRC. Meanwhile, the recognition rates of DLRC and PLRC methods are 70.88\% and 72.19\%, respectively, while that of the DARG method is only 62.58\%. In contrast, all our new models except for DRA-PE-exp outperform the compared methods. In particular, the recognition accuracy of the DRA-TE-eig method reaches 79.51\%, which exceeds the results of all the remaining methods. The high dimensional SPM features form a space with larger capacity, so that the relatively performance of the TE model is improved significantly.

On the other hand, when the SE-ResNeXt-50 features are used to evaluate these methods, we can see that the recognition accuracy of each method is improved. Because the dimension of the SE-ResNeXt-50 features is 2048, which is smaller than the dimension 3000 of the SPM features, the corresponding training and test times are reduced by different degrees. In the compared methods, DARG obtains the best result of 88.59\%, while the results of DLRC, PLRC and KCH-ISCRC are 85.13\%, 84.97\%, and 80.59\%, respectively. Note that our DRA-PE-eig method achieves the best recognition accuracy of 89.80\%, and its training and test times are only 4.1s and 2.9s, respectively.

\subsection{Performance on Caltech256 Database}\label{sec4.7}
The classification task on Caltech256 is harder than Caltech101 with the increasing of categories. We set the number parameter as $(n_{train}, n_{valid}, n_{test})=(10, 10, 10)$. The average results of 30 experiments are reported in Tab. \ref{tab_Cal256}. As the sample size is greater than the dimensionality on Caltech256, we propose the PCA+DRA method to accelerate the training and testing. It first reduces the dimensionality of deep features to 500 by PCA projection. Then complexity of the matrix inverse in ridge regression is reduced to $\mathcal{O}(d^3)$ by Sherman-Morrison-Woodbury formula \cite{golub2013matrix}. For all DRA-based methods, the regression problems are solved in a parallel manner.

\begin{table}[htb]
\caption{Recognition Rate (RR), Standard Error (STE) and CPU Time(s) on Caltech256 Database.}\label{tab_Cal256}
\vspace{-1pc}
\begin{center}
\renewcommand{\tabcolsep}{0.8pc} 
\renewcommand{\arraystretch}{1.3} 
\begin{tabular}{|c|c|c|c|}
\hline
\multicolumn{4}{|c|}{\textbf{SE-ResNeXt-50} ($d=2048$)} \\
\hline
\multirow{2}{*}{Methods} & \multirow{2}{*}{RR $\pm$ STE} & \multicolumn{2}{c|}{Time (s)}  \\
\cline{3-4}
                        &  & Training & Test\\
\hline
AHISD        \cite{HISD}                                     & 87.55$\pm$0.31\% & $-$ & 0.55 \\
CHISD        \cite{HISD}                                     & 87.55$\pm$0.31\% & $-$ & 0.55 \\
DLRC         \cite{DLRC}                                     & 87.51$\pm$0.35\% & $-$ & 0.07 \\
RH-ISCRC     \cite{RISCRC}                                   & 82.01$\pm$0.67\% & 14.10 & 44.99 \\
KCH-ISCRC    \cite{RISCRC}                                   & 81.20$\pm$2.33\% & 29.00 & 1.38 \\
PLRC-\uppercase\expandafter{\romannumeral1}\cite{PLRC}       & 86.25$\pm$0.36\% & $-$ & 9.28 \\
PLRC-\uppercase\expandafter{\romannumeral2}\cite{PLRC}       & 86.94$\pm$0.56\% & $-$ & 5.04 \\
ProCRC        \cite{ProCRC}                                  & 65.27$\pm$0.62\% & $-$ & 0.51 \\
PDL        \cite{wang2017prototype}                          & 83.75$\pm$1.02\% & 33304.14 &  0.03 \\
DARG        \cite{wang2018discriminant}                      & 89.80$\pm$0.52\% & 117.90 & 0.05 \\
NFS                                                          & 88.09$\pm$0.38\% & $-$ & 4.01 \\
\hline
SE-ResNeXt-50 \cite{hu2018squeeze}                          & 62.72$\pm$0.57\% & 621.77 & 0.33 \\
IDLM \cite{shah2016iterative}                               & 77.72$\pm$0.77\% & 500.28 & 0.13 \\
NAN \cite{yang2017neural}                                   & 82.12$\pm$0.58\% & 381.01 & 0.15 \\
\hline
DRA-PE-exp                                                  & 89.33$\pm$0.57\% & 83.59 & 40.55  \\
DRA-TE-exp                                                  & {\bf 91.49$\pm$0.51\%} & 10496.29 & 40.55  \\
DRA-PE-eig                                                  & 89.54$\pm$0.66\% & 41.39 & 40.55  \\
DRA-TE-eig                                                  & 85.32$\pm$0.30\% & 10447.65 & 40.55  \\
\hline
PCA+DRA-PE-exp                                                  & 88.64$\pm$0.53\% & 4.73 & 4.02  \\
PCA+DRA-TE-exp                                                  & 88.42$\pm$0.51\% & 1289.44 & 4.02  \\
PCA+DRA-PE-eig                                                  & 90.35$\pm$0.54\% & 4.08 & 4.02  \\
PCA+DRA-TE-eig                                                  & 91.28$\pm$0.17\% & 1288.83 & 4.02  \\
\hline
\end{tabular}
\end{center}
\end{table}

It is observed that the TE models with stronger criterion outperform the PE models and other methods. DRA-TE-eig achieves the highest accuracy of 91.49\%, while the accuracies of DARG and NAN are 89.80\% and 82.12\%, respectively. The vote rule in SE-ResNeXt-50 and IDLM only obtain accuracies of 62.72\% and 77.72\%, respectively. The developed PCA variants of DRA are really close to the original DRA in performance and also achieve the second-highest accuracy of 91.28\%. Moreover, the training and testing costs of PCA variants are much lower than the originals. All of the above results show that the DRA models are better than other discriminant analysis methods, i.e., DARG and PDL, in most of the cases. It also demonstrates the superiority of residual discriminant learning compared with the discriminant analysis on original data space.

\section{Conclusion}\label{sec5}

In this paper, we propose a discriminant residual analysis method to tackle image set recognition problem with posture and human age variations. DRA attempts to learn the \textit{distance of interest} and then extract discriminant features from \textit{residual} analysis during the training stage. Then it projects the training set and test set into the discriminant subspace simultaneously. With such discriminant projection, the classification results will be more accurate and reliable. Moreover, by using NFS to construct the unrelated groups, DRA is more stable and powerful in real practice. Different regularization strategies are also used to deal with the small sample size problem. Extensive experiment results demonstrate the effectiveness of the proposed methods.

How to extend the discriminant residual analysis method to deal with the zero-shot image classification problem is our future work.

\bibliographystyle{IEEEtran}
\bibliography{egbib}

\end{document}